\documentclass[sigconf]{acmart}

\usepackage{amsmath}
\usepackage{amsfonts}
\usepackage{color}
\usepackage{graphicx}
\usepackage[ruled,linesnumbered]{algorithm2e}

\hfuzz=\maxdimen
\definecolor{DeepGreen}{rgb}{0.00, 0.39, 0.00}
\newcommand \thcom[1]{\textcolor{black}{#1}}

\newcommand \doublecheck[1]{\textcolor{black}{#1}}

\begin{document}

\copyrightyear{2019} 
\acmYear{2019} 
\setcopyright{acmcopyright}
\acmConference[GECCO '19]{Genetic and Evolutionary Computation Conference}{July 13--17, 2019}{Prague, Czech Republic}
\acmBooktitle{Genetic and Evolutionary Computation Conference (GECCO '19), July 13--17, 2019, Prague, Czech Republic}
\acmPrice{15.00}
\acmDOI{10.1145/3321707.3321715}
\acmISBN{978-1-4503-6111-8/19/07}

\title{Algorithm Portfolio for Individual-based Surrogate-Assisted Evolutionary Algorithms}

\author{Hao Tong, Jialin Liu, XinYao}
\authornote{Corresponding author}
\affiliation{%
  \institution{Shenzhen Key Laboratory of Computational Intelligence}
  \institution{University Key Laboratory of Evolving Intelligent Systems of Guangdong Province}
  \institution{Department of Computer Science and Engineering}
  \institution{Southern University of Science and Technology}
  \city{Shenzhen}
  \state{China}
}
\email{ htong6@outlook.com, {liujl, xiny}@sustech.edu.cn }



\begin{abstract}

\doublecheck{Surrogate-assisted evolutionary algorithms (SAEAs) are powerful optimisation tools for computationally expensive problems (CEPs). However, a randomly selected algorithm may fail in solving unknown problems due to no free lunch theorems, and it will cause more computational resource if we re-run the algorithm or try other algorithms to get a much solution, which is more serious in CEPs. In this paper, we consider an algorithm portfolio for SAEAs to reduce the risk of choosing an inappropriate algorithm for CEPs. We propose two portfolio frameworks for very expensive problems in which the maximal number of fitness evaluations is only 5 times of the problem's dimension. One framework named Par-IBSAEA runs all algorithm candidates in parallel and a more sophisticated framework named UCB-IBSAEA employs the Upper Confidence Bound (UCB) policy from reinforcement learning to help select the most appropriate algorithm at each iteration. An effective reward definition is proposed for the UCB policy. We consider three state-of-the-art individual-based SAEAs on different problems and compare them to the portfolios built from their instances on several benchmark problems given limited computation budgets. Our experimental studies demonstrate that our proposed portfolio frameworks significantly outperform any single algorithm on the set of benchmark problems.}

\end{abstract}

\keywords{Algorithm portfolio, surrogate model, evolutionary algorithm, computationally expensive problem}

\maketitle

\section{Introduction}
Computationally expensive problems (CEPs) are very common in many real-world systems, requiring enormous computational resources to accomplish one fitness evaluation \cite{jin2018data}. For instance, one evaluation based on computational fluid dynamic simulations will cost several hours \cite{jin2009systems}. 
Obviously, canonical evolutionary algorithms are challenging to handle this kind of problems directly. To overcome this challenge, surrogate-assisted evolutionary algorithms (SAEAs) are developed by applying a much cheaper model to replace the actual expensive fitness evaluation process to reduce the computational cost \cite{jin2011surrogate}.

Over the past decades, many efficient SAEAs has been proposed and applied into complex real-world applications, such as trauma system \cite{wang2016data}.
Individual-based model control \cite{jin2005comprehensive} method is the most effective strategy that a few individuals will be re-evaluated by the actual function in each generation according to different criteria. For example, some criteria like expected improvements (EI), consider the fidelity of surrogate models and the quality of evaluated solutions simultaneously in global optimisation \cite{jones1998efficient}. On the other hand, recent works in individual-based SAEAs proposed new strategies to trade-off exploration and exploitation of optimisation, like active learning based model management \cite{wang2017committee} and Voronoi-based SAEA framework for very expensive problems \cite{hao2018voronoi}.
They obtain a better solution in test problems and real-world applications compared with classical efficient global optimisation (EGO) \cite{jones1998efficient}.

Even though many model management strategies in individual-based SAEAs are successful in the literature, no free lunch theorems indicate that there is no one best approach appropriate for every problem \cite{wolpert1997no}. For example, EGO is much more powerful than state-of-the-art algorithms in low-dimension cases and algorithm in \cite{yu2019generation} is an expert in multi-modal expensive problems while Voronoi based SAEA framework is good at uni-modal problems \cite{hao2018voronoi}. However, it is hard to determine the optimal algorithm for an unknown problem in practice. In order to address this challenge, algorithm portfolio is employed to reduce the risk of failing to optimise problems in multiple scenarios\cite{huberman1997economics}. \doublecheck{For example, the algorithm portfolio obtained an excellent performance in SAT problems \cite{xu2008satzilla} and noisy optimization problem \cite{cauwet2014algorithm, cauwet2016algorithm}. Although many portfolio strategies have been proposed for noise free evolutionary algorithms \cite{jasper2009self,peng2010population, yuen2016algorithm, baudivs2014online} and Bayesian optimisation \cite{hoffman2011portfolio}, to the best of our knowledge, we are the first to apply the algorithm portfolio for individual-based SAEAs which has shown outstanding performance in solving CEPs.}

We proposed two algorithm portfolio frameworks in this paper for individual-based SAEAs in very expensive problems \cite{hao2018voronoi}. The first framework is motivated from the population-based algorithm portfolio \cite{peng2010population}, which runs all algorithm candidates simultaneously. In another framework, we employ the technique from reinforcement learning to select relatively ``best'' algorithm for every generation. Unlike the portfolio for Bayesian optimisation, we directly choose a method to search a solution for re-evaluation instead of generating several solutions simultaneously by different approaches and then evaluating one of them by actual fitness function \cite{shahriari2016taking}.

The remainder of this paper is structured as follows. Section \ref{related-work} will present some related work about algorithm portfolio. And then the detail of two algorithm portfolio frameworks will be introduced in Section \ref{algorithm portfolio}. In Section \ref{experiments}, we will apply some state-of-the-art individual-based SAEAs to the proposed frameworks and test them in a series of benchmark problems. Finally, the paper will end with a brief conclusion and a discussion of future work in Section \ref{conclusion}.

\section{Related work}\label{related-work}
\subsection{Portfolio of evolutionary algorithm}
In the areas of evolutionary algorithms, algorithm portfolio is applied to increase the probability of finding a better solution by allocating computational resources to several complementary algorithms. The algorithm portfolio frameworks in the literature can be classified into two categories as the parallel-based framework and the sequential-based framework. 



For the parallel-based framework, all candidates will run simultaneously in multiple sub-processes. Population-based algorithm portfolio (PAP) is a typical example \cite{peng2010population}, which allocates computational resources before the optimization according to the prior knowledge. Each algorithm has its own population and evolve independently, but the information is shared among different algorithms by migration strategy. Besides, other parallel-based portfolio frameworks like AMALGAM-SO \cite{jasper2009self} and the UMOEAs \cite{saber2014testing} collect the performance of algorithms during the optimisation process and allocate more resources to the better algorithm. 

On the other hand, the sequential based framework only runs one algorithm at most of the time during the process of optimisation. Different from the parallel-based algorithm portfolio, this kind of framework try to select the best algorithm in different optimisation stage. The multiple evolutionary algorithm (MultiEA) is one of the state-of-the-art sequential algorithm portfolio frameworks \cite{yuen2016algorithm}. It utilises the history convergence curve of each algorithm to predict its performance in the near future, and then the best algorithm will be selected to optimise the problem.  

Another typical sequential portfolio strategy worthy of mention is an online racing algorithm, max-race portfolio (MRP) \cite{tian2014online}. The best algorithm is selected by a statistical test on algorithms' online performance and when enough statistical evidence indicates that one algorithm is significantly inferior to other algorithms, the worst one will be removed by framework permanently.

\subsection{Multi-armed bandit problem}
\doublecheck{The sequential portfolio framework aims at selecting the best optimisation algorithm in the next generation according to the previous performance. The computational resources are dynamically allocated to algorithm constituents. It is similar to the multi-armed bandit framework in reinforcement learning, taking actions to maximise the cumulative reward within limited cost \cite{michael1987bandit, peter2002finite}. }

For a $K$-armed bandit problem, it is basically defined by random variables $\{X_{i, t}|i=1,2,...,K, t\in \mathbb{N}$\} where each $X_{i, t}$ represents an independent and identical distribution with an unknown expectation $\mu_i$ for each arm of bandit machine in $t_{th}$ successful pull \cite{peter2002finite}. For any environment state, the action being taken at next time step is determined by a bandit policy $\pi$, which is learned according to the actions' history rewards. The quality of a policy is measured by cumulative regret, which could be defined by Eq. \eqref{regret}:

\begin{equation}
  R_n = \mu^* n - \sum_{j=1}^{n} E [T_j(n)\mu_j]
  \label{regret}
\end{equation}
where $\mu_j$ is the expected reward of arm $j$, $\mu^*$ is the expectation reward of optimal arm, i.e. $\mu^* \overset{\underset{\mathrm{def}}{}}{=} \max\limits_{1\leq j \leq K}\mu_j$ and $T_j(n)$ represents the number of times arm $j$ has been pulled over $n$ trails. 

The upper confidence bound (UCB) algorithm is a prevalent and effective method for multi-armed bandit problems to tackle the dilemma between exploitation and exploration \cite{peter2002finite}. In this paper, the UCB-Tuned (UCB-t) algorithm is applied for algorithm portfolio because there is no additional parameter requiring to be adjusted in the algorithm which is presented in Eq. \eqref{ucb-t}:

\begin{equation}
  \pi_{j, n} = \overline {{\mu}} _j  + \sqrt {\frac{{\ln {\rm{n}}}}{{{T_j(n)}}} \cdot \min \{ \frac{1}{4},{v_j}({T_j(n)})\} }
  \label{ucb-t}
\end{equation}
and
\begin{equation}
{v_j}(s) = \frac{1}{s}\mathop \sum \limits_{\tau  = 1}^s \mu _{j,\tau }^2 - \overline \mu  _{j,s}^2 + \sqrt {\frac{{2\ln n}}{s}}
\label{ucb-t1}
\end{equation}
where $\bar{\mu}_j$ is the average reward of arm $j$ after $n$ trails. The policy will select the arm with maximal UCB value according to Eq. \eqref{ucb-t} and \eqref{ucb-t1} for the next generation. 

In the literature, there has been some works about bandit framework for algorithm selection. Baudi{\v{s}} and Po{\v{s}}{\'\i}k~\cite{baudivs2014online} applied basic UCB in black box optimisation in which they defined the reward by introducing a log-rescaling method to process the raw fitness value. And the value rank \cite{fialho2010toward} as a method of reward definition is also compared in the experiments. The results show that the UCB algorithm is efficient in algorithm selection problems. Also in \cite{david2014differential}, authors regarded the algorithm selection problem as a non-stationary bandit problem and applied UCB algorithm to be the decision policy.

From this view, it is reasonable to consider the algorithm portfolio problem in the area of reinforcement learning and employ appropriate methods to construct the framework for individual-based SAEAs.



\section{Algorithm portfolio strategies} \label{algorithm portfolio}
Individual-based SAEAs re-evaluate a few individuals at each generation and individuals being re-evaluated in the next generation is only determined by the current database. As a sequence, we will introduce two portfolio frameworks as parallel individual-based SAEAs and UCB for individual-based SAEAs which are motivated from two different aspects as reviewed previously. 

\subsection{Parallel individual-based SAEAs}

\begin{figure}[!htpb]
  \includegraphics[scale=0.65]{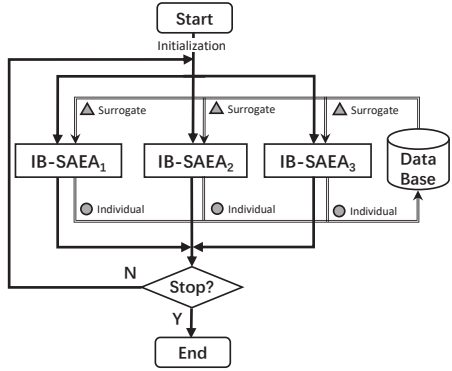}
  \caption{The diagram of the framework: Parallel individual-based SAEAs.}
  \label{par-ibsaea}
\end{figure}

Similar with the algorithm portfolio for canonical evolutionary algorithms, it is intuitive to consider each individual-based SAEA as a simple evolutionary algorithm and embed them into the existing framework, like PAP or MultiEA. From this aspect, the parallel individual-based SAEAs (Par-IBSAEA) framework is proposed that all algorithm candidates run simultaneously at each generation. Nevertheless, it is more convenient than portfolio for canonical evolutionary algorithms because almost all individual-based SAEAs have the same algorithm structure and it does not require a particular design for each algorithm. A brief diagram for Par-IBSAEA with only three algorithm instances is presented in Figure \ref{par-ibsaea} where the double solid line represents the interaction between individual-based SAEAs and the database.

Considering the set of individual-based SAEAs $\mathcal{A} = \{A_1, A_2, ..., A_n\}$ with $n$ algorithms, each algorithm is independent with each other. The framework starts by the initialization of a database and all SAEAs use the same surrogate model constructed by samples in the database to search for the next re-evaluated solution. Due to different mechanisms for various algorithms, they will obtain several different solutions for re-evaluation as $\mathcal{D} = \{\mathbf{x}_1, \mathbf{x}_2,..., \mathbf{x}_k\}$, where the size of $k$ is not completely equal to $n$ because some individual-based SAEAs might get more than one solution. For example, VESAEA will re-evaluate two solutions in one generation when the optimisation in global search stage \cite{hao2018voronoi}. The database will be updated by adding evaluated solutions at the end of one iteration, and the whole algorithm will be stopped when the fitness evaluations are exhausted.


\subsection{UCB for individual-based SAEAs}
\begin{figure}[!htpb]
  \includegraphics[scale=0.65]{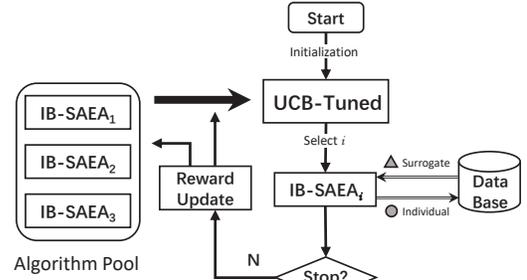}
  \caption{The diagram of the framework: UCB for individual-based SAEAs.}
  \label{ucb-ibsaea}
\end{figure}
On the other hand, the individual-based SAEA portfolio is regarded as a multi-armed bandit problem that an individual-based SAEA is considered as an arm and the quality of solutions could be used to measure the reward of actions. Then, we could apply the UCB algorithm to determine which algorithm is used in the next generation for CEPs.

The key point in UCB for individual-based SAEAs (UCB-IBSAEA) is the definition of reward. According to the condition of UCB, the reward used in UCB policy must be in the range of $[0, 1]$. \thcom{It is intuitive to define the reward of an algorithm by linear scaling the solution's fitness into the bound of problem.} However, we can not obtain the actual bounds due to the problem solved is a black box. And the only information we could use is the evaluated individuals in the database. Therefore, the bounds could be estimated by the samples' fitness in the database. Assuming all solutions' fitness in database in $t_{th}$ generation is $\mathcal{Y}_t = \{y_1, y_2, ..., y_n\}$, the empirical upper bound $eUB_{t}$ and empirical lower bound $eLB_{t}$ could be estimated by Eq. \eqref{ebound}:
\begin{equation}
  \left\{\begin{matrix}
    eUB_t = \max(\mathcal{Y}_t)\\
    eLB_t = \min(\mathcal{Y}_t)    
    \end{matrix}\right.
  \label{ebound}
\end{equation}
and the empirical upper bound and lower bound will be updated after updating the database. As a sequence, the reward of one algorithm could be formulated as Eq. \eqref{reward}:
\begin{equation}
  \mu_{j, t} = \frac{eUB_{t}-f_{i, t}}{eUB_t-eLB_t}
  \label{reward}
\end{equation}
where $f_{i,t}$ is the actual fitness of solution found by the algorithm in $t_{th}$ iteration. 

However, the performance of empirical bound estimated by Eq. \eqref{ebound} will be bad due to the characteristic of the optimisation process. It is obvious that the convergence speed varies with the optimisation stage that the speed is much faster at the beginning and slows down when the optimization is near convergence. \doublecheck{Therefore, the empirical bounds are almost fixed and evaluated solutions' fitness are almost all very close to the empirical lower bound in the later stage of optimisation. As a sequence, the reward between different algorithms almost has no significant difference and the UCB policy is not effective any more. Hence, the simple normalization of fitness to be the reward is not appropriate for the UCB policy.}

In order to improve the stability of the UCB policy, we introduce an online strategy to update the estimated bound by sliding window. The main idea is shown in Algorithm \ref{update-bound}. Considering all evaluated solutions in the database after $t_{th}$ generation, a sliding window with the size of $sw$ selects best $sw$ solutions to form a subset. The $sw$ is set as $2d$ in this work, where $d$ is the dimension of a problem. If the minimal fitness in the subset is lower than the current $eLB$ which indicates the problem's lower bound could be a much lower value, the $eLB$ will be updated by the newest minimal value. And if the maximal fitness in the subset is lower than current $eUB$ which means the optimisation has entered into another stage compared to the last stage, $eUB$ will be updated to refine the empirical bounds.

\begin{algorithm}
  \caption{Update empirical bounds by sliding window.}\label{update-bound}
  \KwIn{Database, lower bound $eLB_t$, upper bound $eUB_t$ }
  Size of sliding window: $sw$ \\
  Sort evaluated samples in database from best to worst; \\
  The top $sw$ best individuals $\mathcal{S} = \{(\mathbf{x}_1, y_1), ..., (\mathbf{x}_{sw}, y_{sw})\}$; \\
  Collect fitness of individuals in $\mathcal{S}$: $\mathcal{Y} = [y_1, y_2, ..., y_{sw}]$; \\
  \eIf {$eLB_t > \min(\mathcal{Y})$}
  {
    $eLB_{t+1} = \min(\mathcal{Y})$
  }
  {
    $eLB_{t+1} = eLB_t $
  }

  \eIf {$eUB_t > \max(\mathcal{Y})$}
  {
    $eUB_{t+1} = \max(\mathcal{Y})$
  }
  {
    $eUB_{t+1} = eUB_t $
  }
  \KwOut{New emprical bound: $eLB_{t+1}, eUB_{t+1}$}
\end{algorithm}

After defining the algorithm's reward, the UCB-IBSAEA framework is easy to be implemented which is presented in the Figure \ref{ucb-ibsaea}. The UCB-t with no additional parameter in Eq. \eqref{ucb-t} is used in the framework. The framework starts by initialising a database and then the UCB-IBSAEA will select one best algorithm $A_k$ from the algorithm pool $\mathcal{A} = \{A_1, A_2, ..., A_n\}$ according to the UCB-t policy for the following generation. The new solution will be added into the database after being evaluated by the actual fitness function. After that, the empirical bounds and reward information for each algorithm will be updated. Finally, the optimisation process will stop after running out of all fitness evaluations.

\section{Numerical experiments and analysis} \label{experiments}
The performance of two proposed portfolio frameworks is studied in this section. We choose three state-of-the-art individual-based SAEAs to generate two portfolio instances and evaluate them on a set of benchmark functions. 

\subsection{Experimental setting}
In this work, very computationally expensive problems are considered so that we employ three efficient individual-based SAEAs to asses the efficacy of portfolio frameworks. Three algorithms are expert in different kinds of problems, and a brief introduction for them is presented below: 
\begin{itemize}
  \item EGO-LCB: Efficient global optimisation with lower confidence bound \cite{jones1998efficient} is a very efficient algorithm for low-dimension expensive problems. 
  \item VESAEA: Voronoi-based efficient SAEA employed Voronoi diagram in SAEA framework to assist the local search process \cite{hao2018voronoi}. It is good at uni-modal problems even though there are some noises over the landscape.  
  \item GORS-SSLPSO: \thcom{Despite it is a so-called generation based SAEA \cite{yu2019generation}, we consider it as the individual based SAEA because the solution re-evaluated in next generation is determined by the current database. The algorithm maintains the diversity by PSO operator and guarantees the convergence by a restart strategy. And it is appropriate for solving multi-modal problems.}
\end{itemize}

\begin{table}[htpb!]\tiny
  \centering
  \caption{Benchmark Problems. The optimum solution is shifted to another random position in the landscape.}\label{test-problems}
  \resizebox{\linewidth}{!}{
  \begin{tabular}{|c|c|c|c|}
  \hline
  Problem    & Dimension     & Optimum & Note                                                                       \\ \hline
  Sphere     & 10,20,30      & 0       & Uni-modal                                                                  \\ \hline
  Griewank   & 10,20,30      & 0       & Multi-modal                                                                \\ \hline
  Ackley     & 10,20,30      & 0       & Multi-modal                                                                \\ \hline
  Rosenbrock & 10,20,30      & 0       & Multi-modal with narrow valley                                             \\ \hline
  Rastrigin  & 10,20,30      & 0       & Very complicated multi-modal                                               \\ \hline
  \end{tabular}}
  \end{table}
  
Frameworks are tested on five widely used problems with dimensions $d = 10, 20, 30$ as shown in Table \ref{test-problems}. The maximal fitness evaluation is setting as $5d$ where $2d$ is used for initialisation, and all comparisons are based on 25 independent runs. In the following, the comparison between portfolio frameworks and three single SAEAs will be performed firstly and then we will analyse the performance of portfolio frameworks by comparing with another two frameworks: random selection framework and epsilon-greedy selection framework \cite{peter2014coco}.

\subsection{Comparative results with single algorithms}
\begin{table*}[!htpb]
  \caption{The averaged best results and standard deviation of algorithm portfolio and single algorithm on test problems with 25 independent runs. The `win-draw-lose' represents the control method is superior, not significantly different and inferior to the compared algorithm.} \label{result1}  
\small 
\begin{tabular}{|c|c|c|c|c|c|c|} \hline Problem & D &  UCB-IBSAEA  & Par-IBSAEA & GORS-SSLPSO & VESAEA & EGO-LCB \\ \hline       sphere & 10  & 1421.21 $\pm$ 939.89  & \bf{1039.95 $\pm$ 712.93}  & 5483.17 $\pm$ 2789.23  & 1878.56 $\pm$ 691.56  & 5728.52 $\pm$ 2791.72  \\ \hline       sphere & 20  & 2445.62 $\pm$ 1109.81  & \bf{2417.95 $\pm$ 938.53}  & 9604.73 $\pm$ 2911.24  & 10906.42 $\pm$ 1487.33  & 44083.71 $\pm$ 7045.51  \\ \hline       sphere & 30  & 3606.77 $\pm$ 1377.47  & \bf{3091.30 $\pm$ 1171.53}  & 15133.57 $\pm$ 4323.70  & 25736.52 $\pm$ 2554.90  & 81185.30 $\pm$ 8810.83  \\ \hline       rosenbrock & 10  & 209.05 $\pm$ 82.43  & 234.90 $\pm$ 112.40  & \bf{89.03 $\pm$ 46.93}  & 238.17 $\pm$ 101.78  & 1127.73 $\pm$ 399.72  \\ \hline       rosenbrock & 20  & 349.14 $\pm$ 105.95  & 512.16 $\pm$ 144.50  & \bf{281.95 $\pm$ 314.65}  & 936.38 $\pm$ 240.64  & 5281.28 $\pm$ 1460.20  \\ \hline       rosenbrock & 30  & 418.05 $\pm$ 86.88  & 596.34 $\pm$ 150.43  & \bf{311.92 $\pm$ 116.62}  & 2419.77 $\pm$ 435.33  & 14763.02 $\pm$ 2247.72  \\ \hline       ackley & 10  & 19.46 $\pm$ 0.69  & 19.11 $\pm$ 1.39  & 19.52 $\pm$ 0.85  & 16.61 $\pm$ 1.89  & \bf{15.71 $\pm$ 5.35}  \\ \hline       ackley & 20  & 19.18 $\pm$ 0.58  & 19.27 $\pm$ 0.70  & 19.49 $\pm$ 0.51  & \bf{18.43 $\pm$ 0.74}  & 20.48 $\pm$ 0.23  \\ \hline       ackley & 30  & \bf{19.06 $\pm$ 0.59}  & 19.22 $\pm$ 0.50  & 19.52 $\pm$ 0.59  & 19.13 $\pm$ 0.37  & 20.75 $\pm$ 0.18  \\ \hline       griewank & 10  & 15.18 $\pm$ 12.21  & \bf{12.82 $\pm$ 9.30}  & 78.07 $\pm$ 37.73  & 17.83 $\pm$ 5.11  & 46.91 $\pm$ 20.39  \\ \hline       griewank & 20  & 27.34 $\pm$ 12.03  & \bf{18.84 $\pm$ 5.62}  & 143.75 $\pm$ 41.18  & 101.19 $\pm$ 11.65  & 386.30 $\pm$ 58.78  \\ \hline       griewank & 30  & 39.46 $\pm$ 13.21  & \bf{31.33 $\pm$ 12.98}  & 209.52 $\pm$ 48.56  & 226.61 $\pm$ 15.51  & 714.12 $\pm$ 92.21  \\ \hline       rastrigin & 10  & 73.63 $\pm$ 16.69  & 79.21 $\pm$ 18.00  & \bf{55.79 $\pm$ 24.19}  & 81.52 $\pm$ 14.91  & 101.07 $\pm$ 20.36  \\ \hline       rastrigin & 20  & 117.38 $\pm$ 30.23  & 152.84 $\pm$ 34.60  & \bf{86.66 $\pm$ 31.20}  & 165.87 $\pm$ 34.84  & 271.99 $\pm$ 19.77  \\ \hline       rastrigin & 30  & 143.47 $\pm$ 29.35  & 176.62 $\pm$ 26.82  & \bf{114.12 $\pm$ 25.37}  & 235.87 $\pm$ 53.24  & 467.43 $\pm$ 28.99  \\ \hline      \multicolumn{2}{|c|}{Average-Ranking}  & 2.07  & 2.20  & 2.73  & 3.33  & 4.67  \\ \hline     \multicolumn{2}{|c|}{Wilcoxon-Test} & Control method  & 4-9-2  & 7-2-6  & 10-3-2  & 14-1-0 \\ \hline \multicolumn{2}{|c|}{Wilcoxon-Test}  & 2-9-4  & Control method & 6-3-6  & 9-4-2  & 14-1-0  \\ \hline  \end{tabular}
\end{table*}

\begin{figure}[htpb!]
  \centering
  
  \includegraphics[width=.45\linewidth]{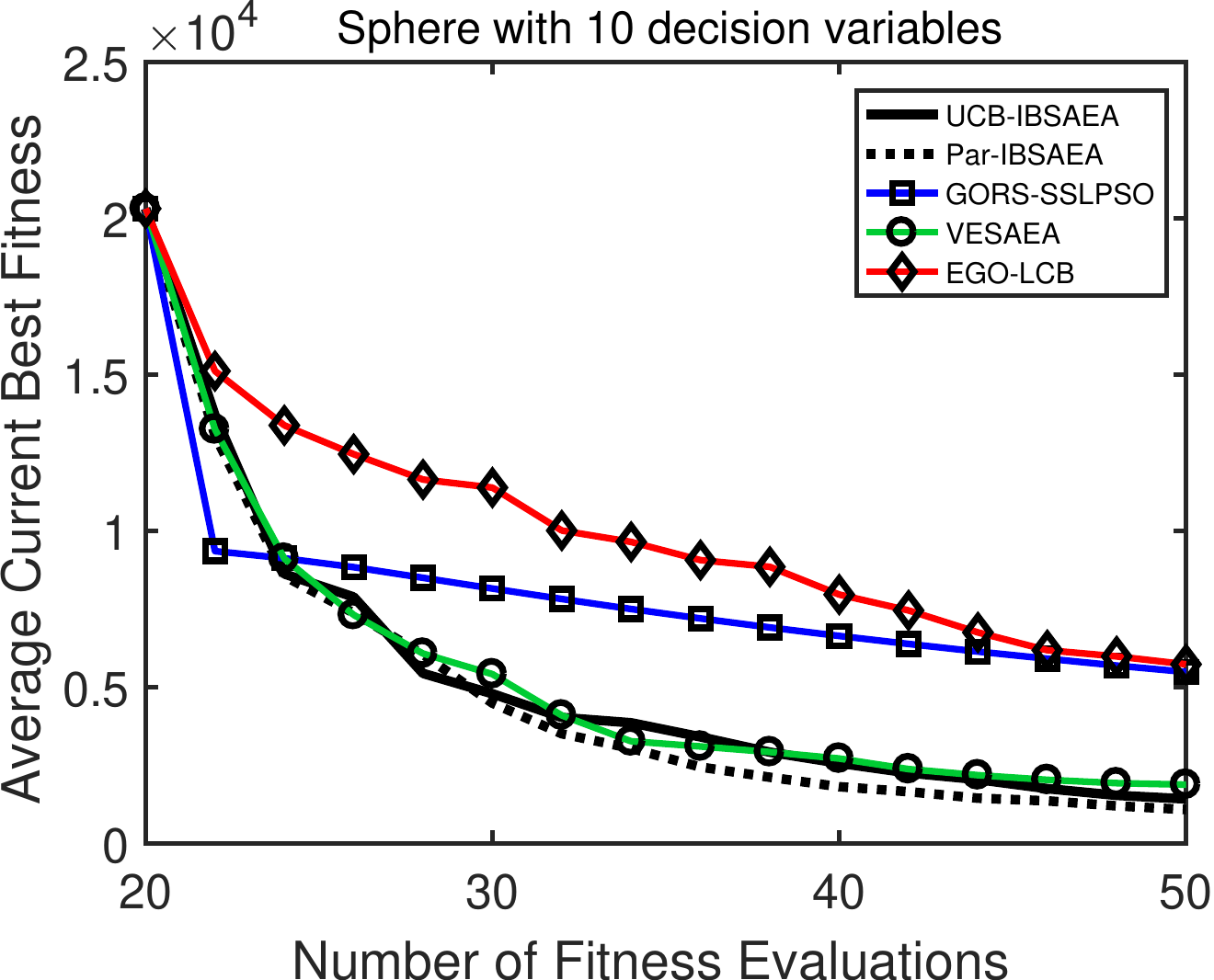}\quad
  \includegraphics[width=.45\linewidth]{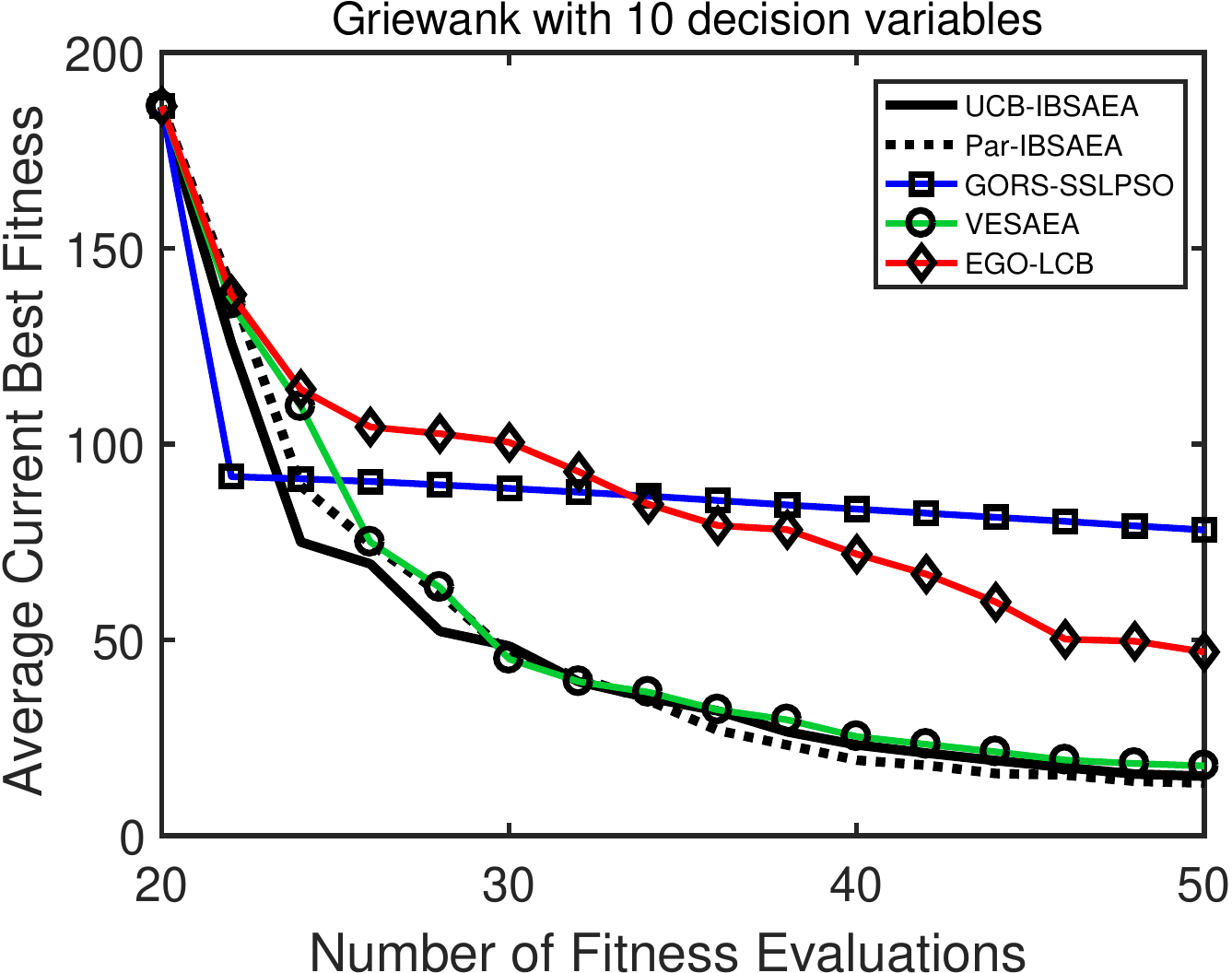}\quad
  \medskip
  
  \includegraphics[width=.45\linewidth]{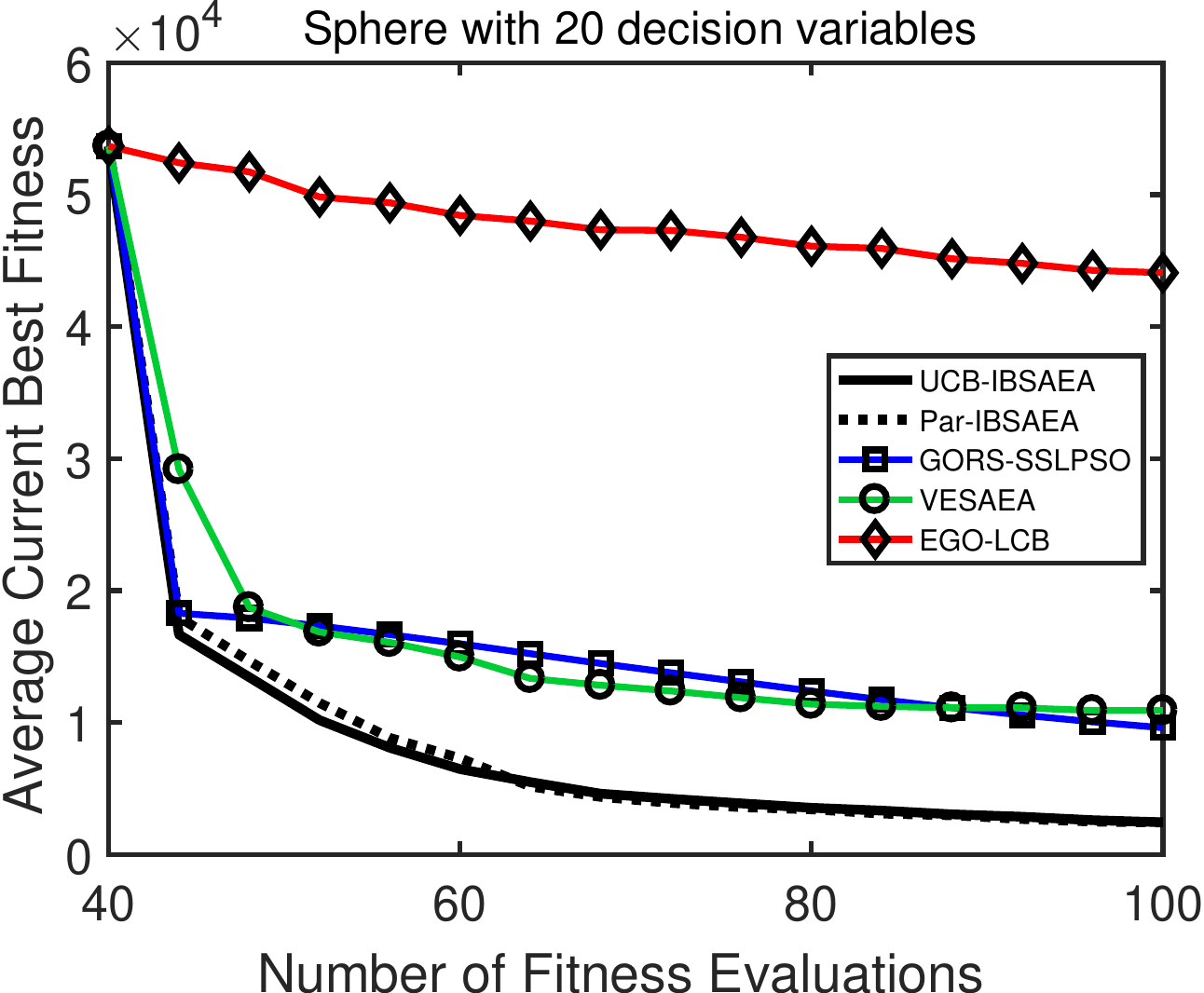}\quad
  \includegraphics[width=.45\linewidth]{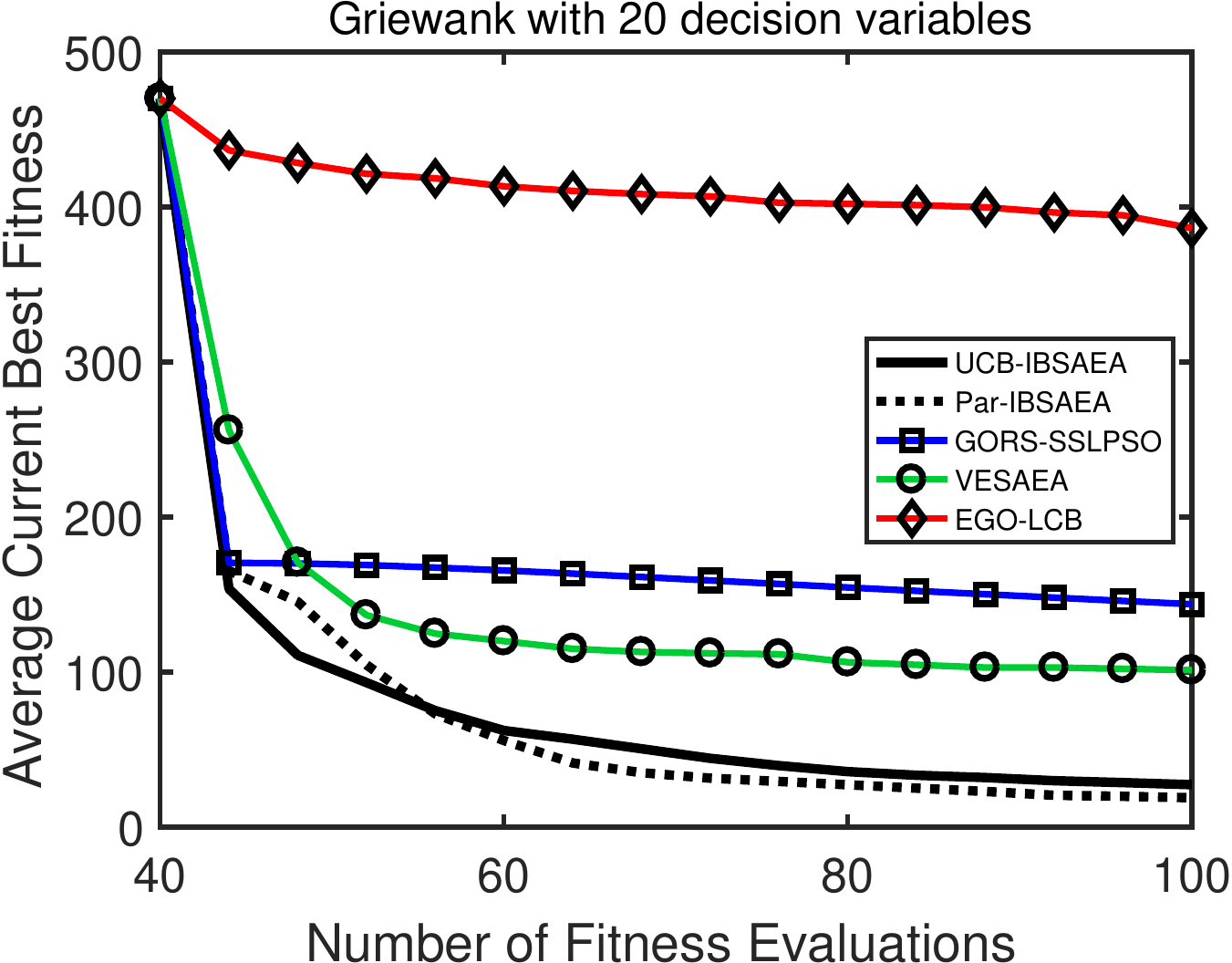}\quad
  \medskip
  
  \includegraphics[width=.45\linewidth]{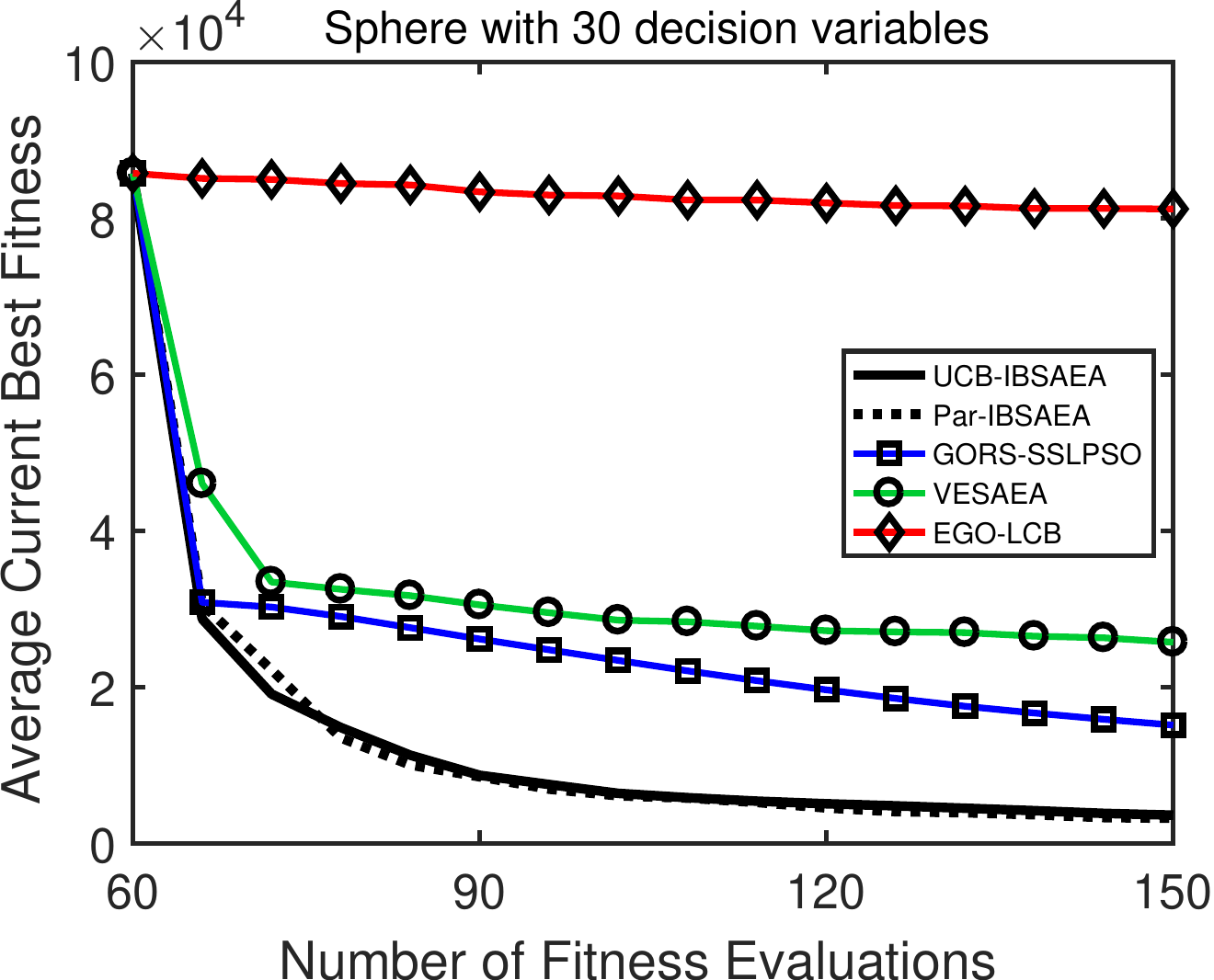}\quad
  \includegraphics[width=.45\linewidth]{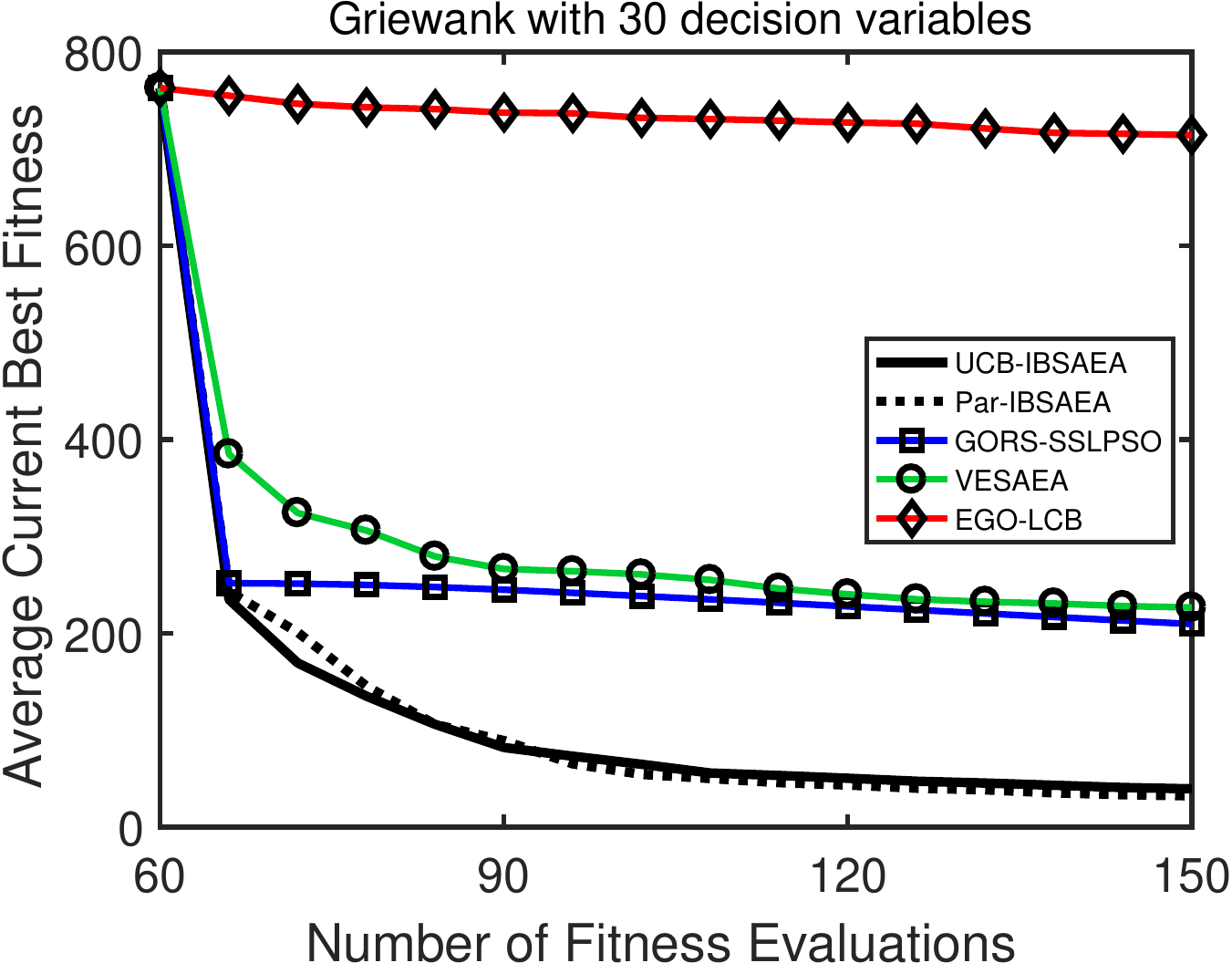}\quad
  \caption{Convergence curves of comparison algorithms on the Sphere and Griewank function with $D = 10,20,30$.}
  \label{sphere-griewank}
  \end{figure}

\begin{figure}[htpb!]
  \centering

  \includegraphics[width=.45\linewidth]{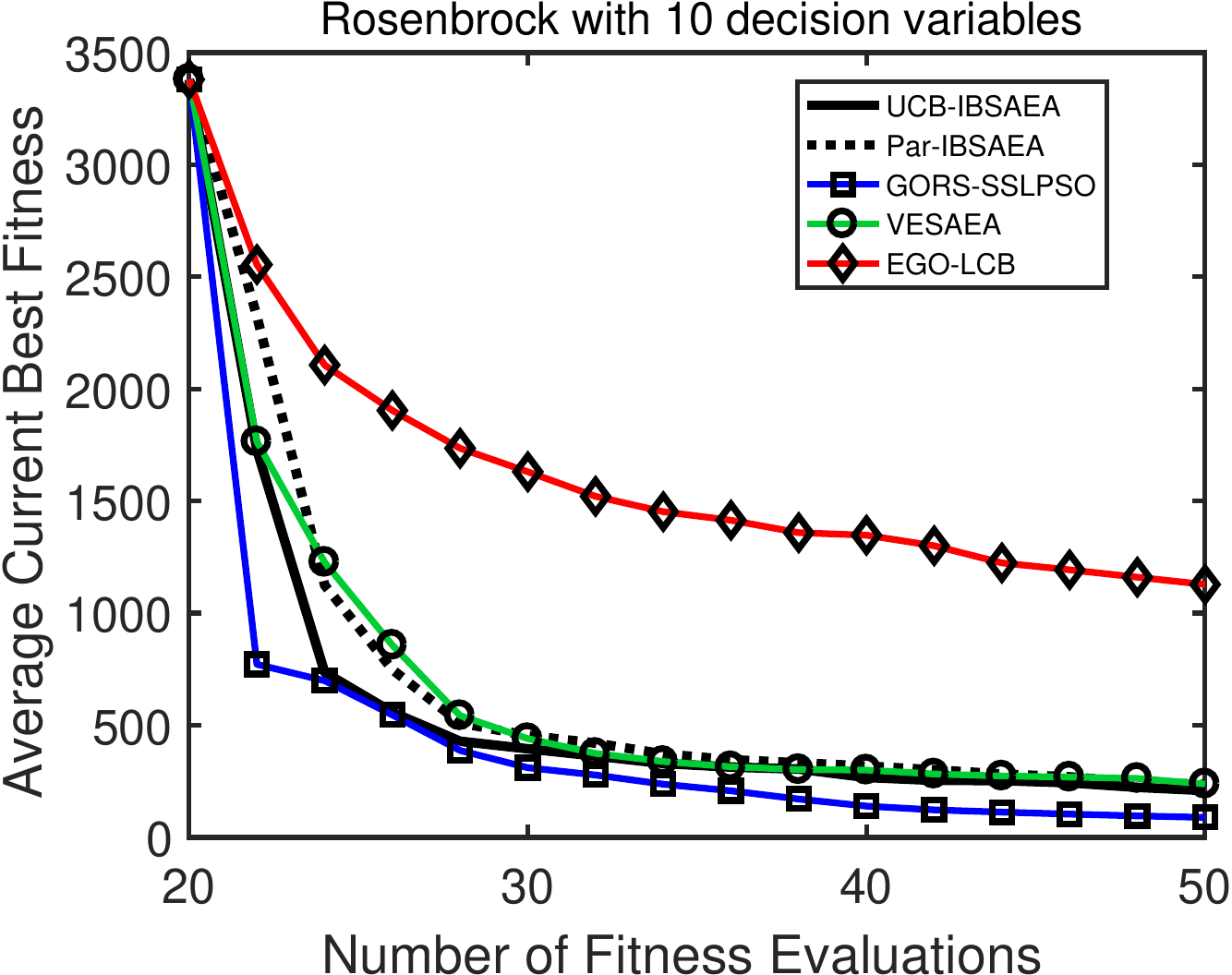}\quad    
  \includegraphics[width=.45\linewidth]{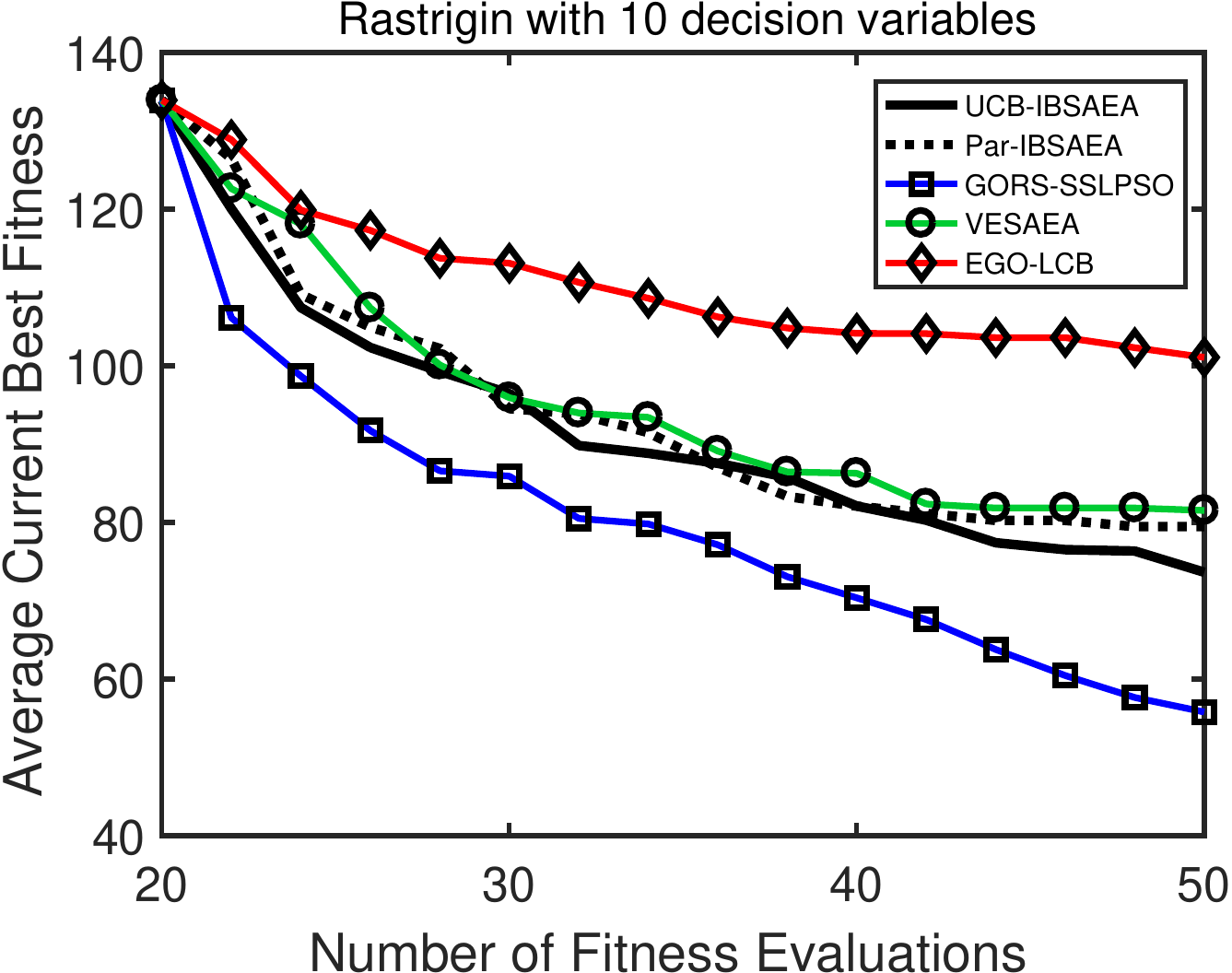}\quad
  \medskip

  \includegraphics[width=.45\linewidth]{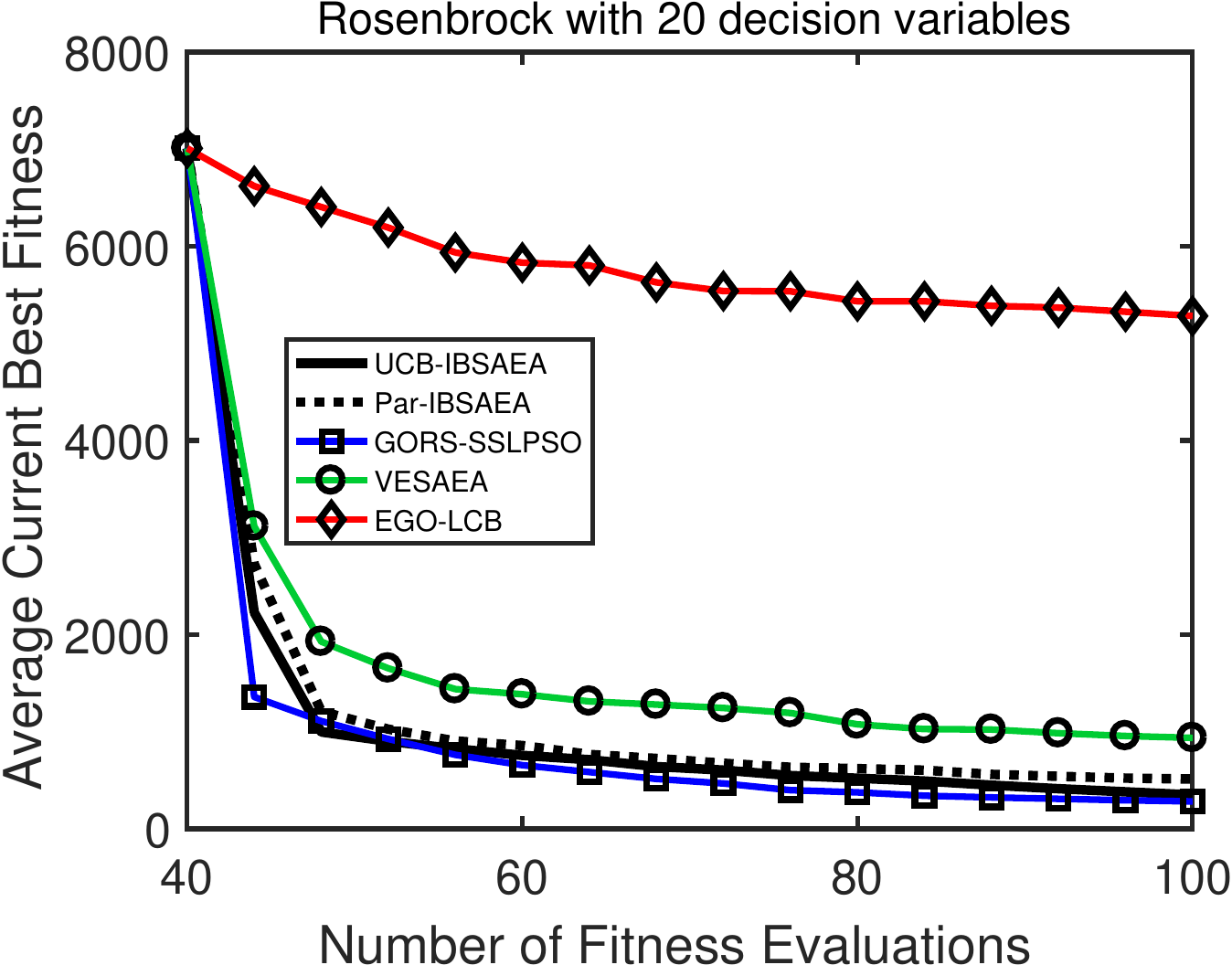}\quad    
  \includegraphics[width=.45\linewidth]{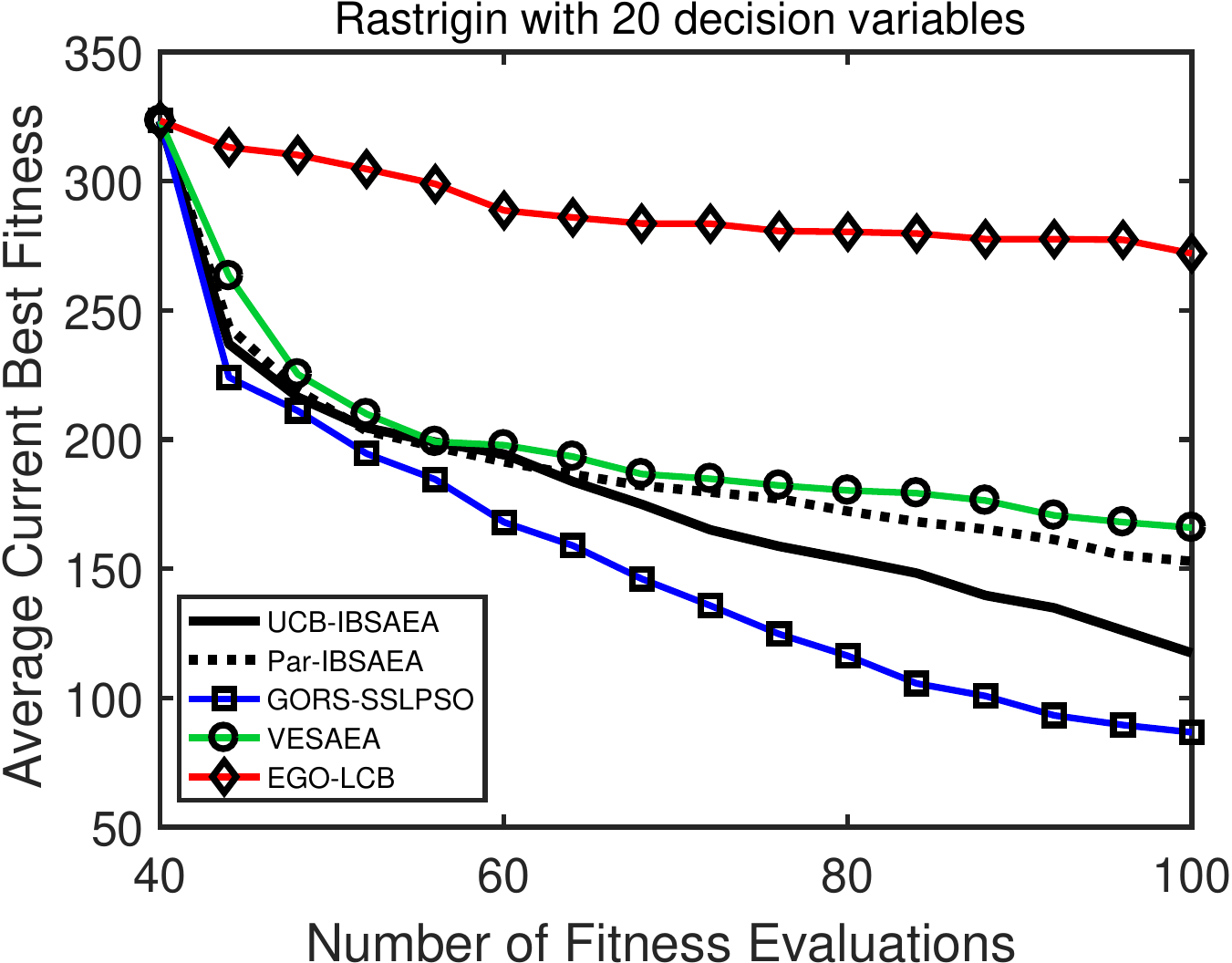}\quad
  \medskip

  \includegraphics[width=.45\linewidth]{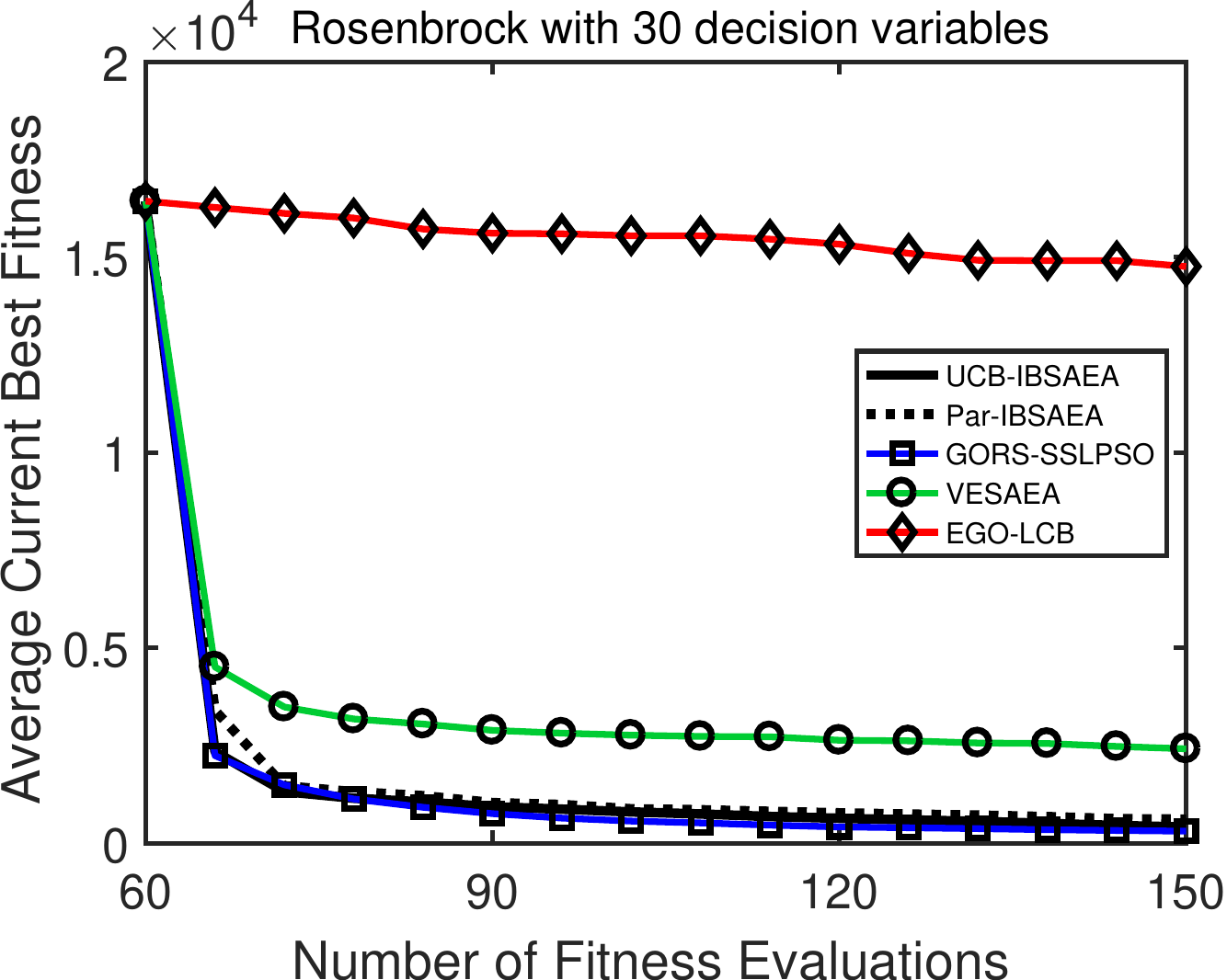}\quad   
  \includegraphics[width=.45\linewidth]{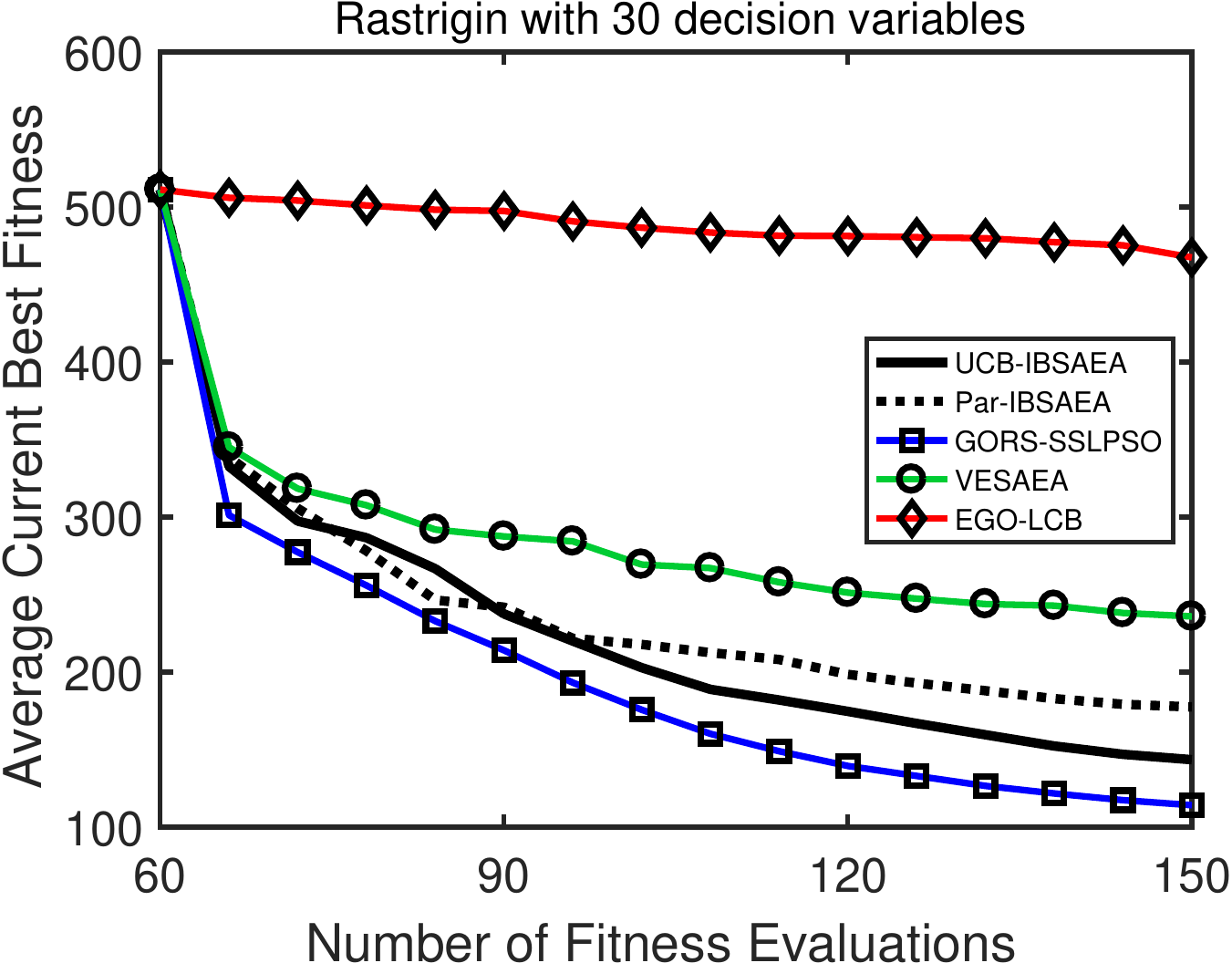}\quad

  \caption{Convergence curves of comparison algorithms on the Rosenbrock and Rastrigin function with $D = 10,20,30$.}
  \label{rastrigin-rosenbrock}
  \end{figure}

\begin{figure}[htpb!]
  \centering
  \includegraphics[width=0.45\linewidth]{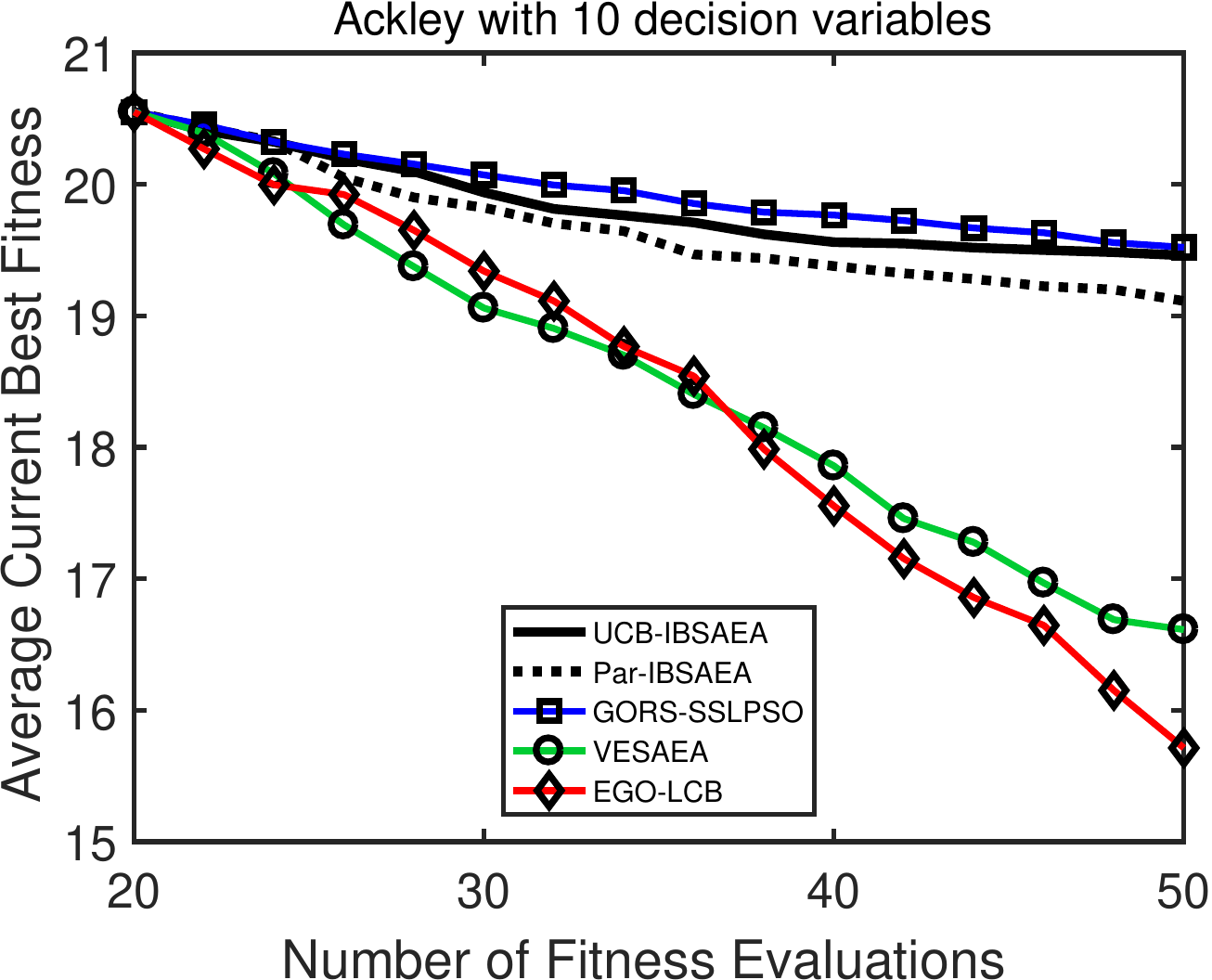}\quad
  \includegraphics[width=0.45\linewidth]{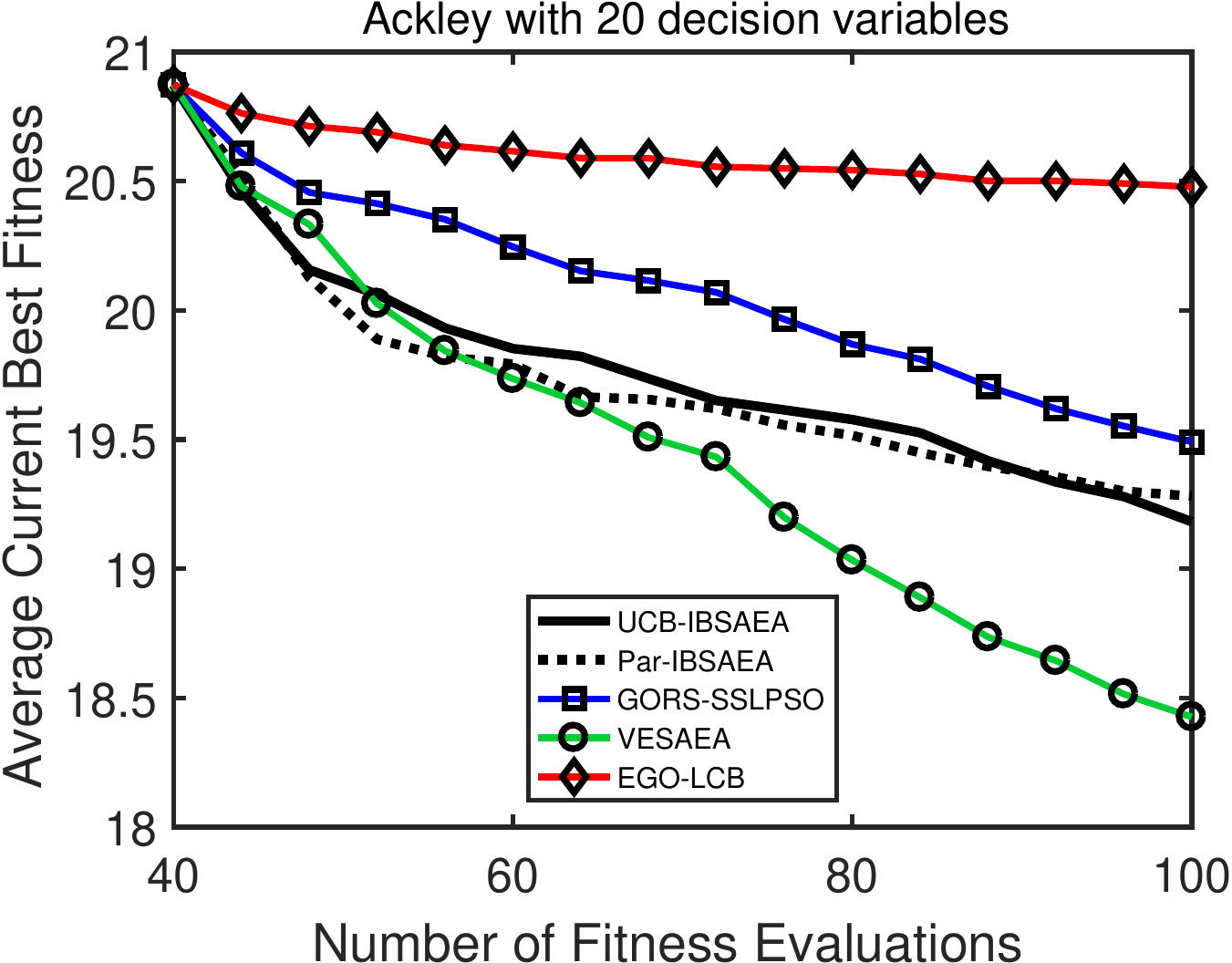}\quad
  
  \includegraphics[width=0.45\linewidth]{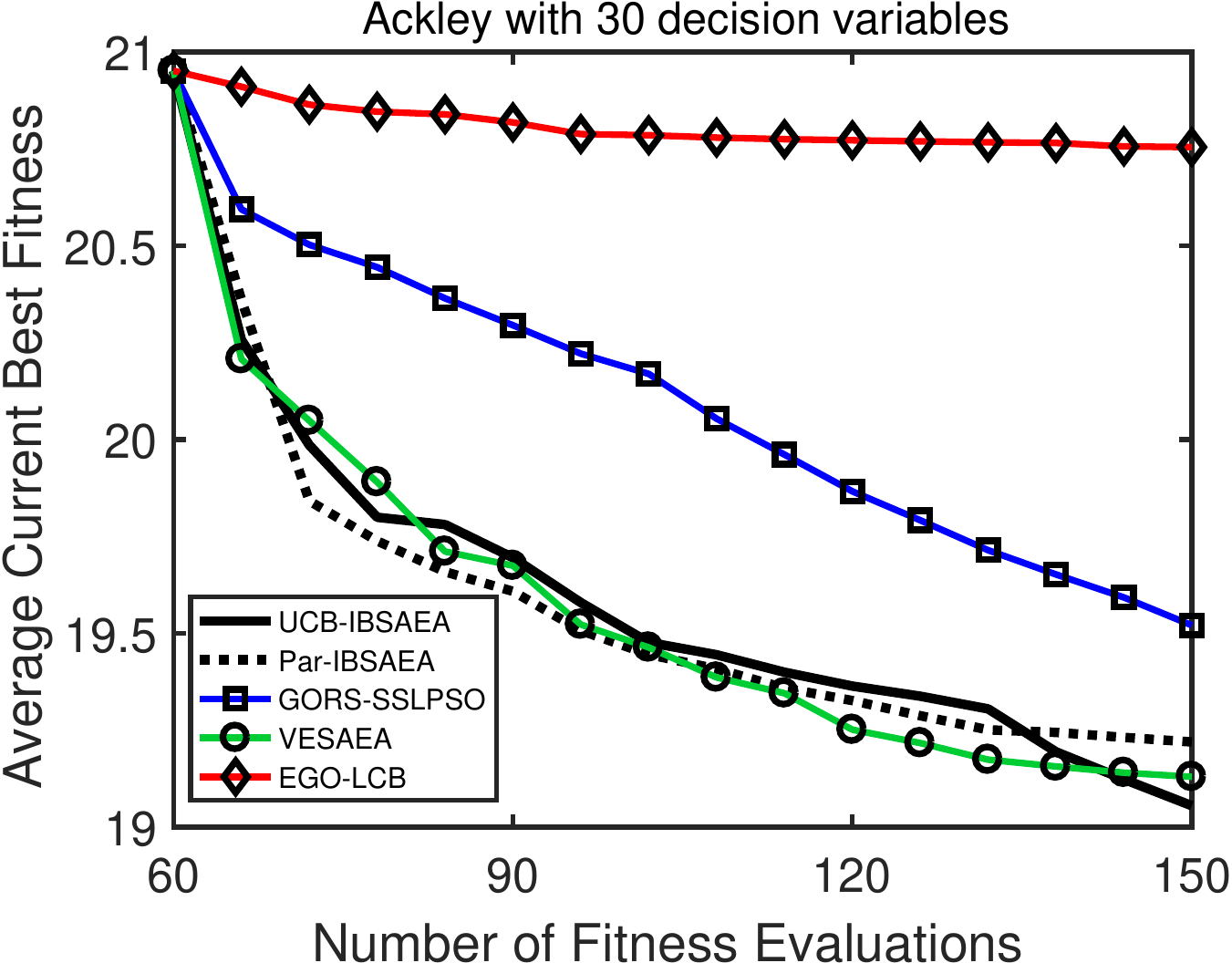}\quad
  

  \caption{Convergence curves of comparison algorithms on the Ackley function with $D = 10,20,30$.}
  \label{ackley}
  \end{figure}

The results of two proposed frameworks and three algorithm candidates on benchmark problems over 25 independent runs are presented in Table \ref{result1} where figures in each cell denote averaged best fitness and standard deviation, in which bold ones are the best results among five algorithms in one problem. The averaged ranking for algorithm portfolio and three SAEAs are listed on the third row from the bottom, from which we could find the proposed frameworks are much better than all single algorithms. Moreover, the last two rows provide the result of Wilcoxon test with a 0.05 significance level, in which the UCB-IBSAEA and Par-IBSAEA are control methods, respectively and the `win-draw-lose' represents the control method is superior, not significantly different and inferior to the compared algorithm. Convergence profiles of all algorithms on 15 test problems are plotted in Figs. \ref{sphere-griewank}-\ref{ackley},  where the x-axis ranges from $2D$ to $5D$ because the first $2D$ fitness evaluations are used for initialization, which is same for all algorithms in each run for fair comparison.

The overall performance of an algorithm portfolio is significantly better than each single algorithm in Table \ref{result1}. However, the performance of an algorithm portfolio in different problems has a significant difference. For example, in sphere and griewank problems, the portfolio dramatically improves the quality of the final solution. By contrast, they are not always the best in rosenbrock and rastrigin problems, but it still can get an acceptable result compared with the best single algorithm. 

The sphere function is a uni-modal problem and griewank function can also be regarded as a uni-modal problem with many small noises in the whole landscape. It is obvious in Table \ref{result1} that the portfolio performs much better than every single algorithm on these kind of problems. \doublecheck{Nonetheless, we could find the portfolio's effect is much higher when the dimension equal to 20 and 30 even though there is a significant terrible algorithm in the framework as shown in Figure \ref{sphere-griewank}.} In 10-dimension case, the portfolio is similar to the performance of VESAEA which is the best algorithm for this kind of problem among three candidates.

Rosenbrock and rastrigin problems are both multi-modal problems which the performance of optimisation algorithms is likely to be hindered by attractive local optimums. GORS-SSLPSO is the most appropriate optimization algorithm for multi-modal problems among three algorithm candidates. And two algorithm portfolio frameworks are a little worse than GORS-SSLPSO, but performs better than other two algorithms as shown in convergence profiles in Figure \ref{rastrigin-rosenbrock}.

In Ackley problems, although the superiority of algorithm portfolio is not obvious, portfolio still obtains relatively better result compared with the worst single algorithm. It is probably due to the little difference between three algorithms for this problem, and the performance of portfolio is restricted by the single algorithm's ability. 

As introduced above, we could obtain the conclusion that two proposed frameworks could improve the performance of single algorithms in uni-modal problems and obtain the similar performance with the most appropriate algorithm candidate in complex multi-modal problems. For uni-modal problems, there is only one attractive global optimal solution over the whole landscape and the algorithm portfolio performs better than every single algorithm probably because various algorithms provide more diversity during the optimization process. VESAEA and EGO-LCB are good at exploiting the search space and GORS-SSLPSO could contribute more diversity so that they promote the performance of each other in uni-modal problems. But in the multi-modal problem which has many attractive local optimums, although the GORS-SSLPSO can provide much diversity, other two algorithms are easy to be trapped in the local optimal. As a result, the whole performance of the portfolio is hindered by VESAEA and EGO-LCB so that the result of the portfolio is slightly worse than the GORS-SSLPSO algorithm.

\subsection{Comparative results with other frameworks}
\begin{table*}[!htpb]
  \caption{The averaged best results and standard deviation of four different frameworks: UCB-IBSAEA, Par-IBSAEA, RS, EG50 on test problems with 25 independent runs. The `win-draw-lose' represents the control method is superior, not significantly different and inferior to the compared algorithm.} \label{result2}  
\small  \begin{tabular}{|c|c|c|c|c|c|} \hline Problem & D &  UCB-IBSAEA  & Par-IBSAEA& RS &  EG50  \\ \hline      sphere & 10  & 1421.21 $\pm$ 939.89  & \bf{1039.95 $\pm$ 712.93}  & 1178.42 $\pm$ 567.00  & 1280.76 $\pm$ 494.39  \\ \hline      sphere & 20  & 2445.62 $\pm$ 1109.81  & \bf{2417.95 $\pm$ 938.53}  & 2598.02 $\pm$ 1144.57  & 4455.35 $\pm$ 1584.86  \\ \hline      sphere & 30  & 3606.77 $\pm$ 1377.47  & \bf{3091.30 $\pm$ 1171.53}  & 3134.88 $\pm$ 995.80  & 4980.92 $\pm$ 1781.29  \\ \hline      rosenbrock & 10  & \bf{209.05 $\pm$ 82.43}  & 234.90 $\pm$ 112.40  & 268.05 $\pm$ 143.15  & 264.17 $\pm$ 108.04  \\ \hline      rosenbrock & 20  & \bf{349.14 $\pm$ 105.95}  & 512.16 $\pm$ 144.50  & 467.78 $\pm$ 174.07  & 561.89 $\pm$ 183.74  \\ \hline      rosenbrock & 30  & \bf{418.05 $\pm$ 86.88}  & 596.34 $\pm$ 150.43  & 604.03 $\pm$ 134.11  & 721.90 $\pm$ 194.39  \\ \hline      ackley & 10  & 19.46 $\pm$ 0.69  & 19.11 $\pm$ 1.39  & \bf{18.66 $\pm$ 1.10}  & 18.82 $\pm$ 1.35  \\ \hline      ackley & 20  & \bf{19.18 $\pm$ 0.58}  & 19.27 $\pm$ 0.70  & 19.36 $\pm$ 0.57  & 19.36 $\pm$ 0.47  \\ \hline      ackley & 30  & \bf{19.06 $\pm$ 0.59}  & 19.22 $\pm$ 0.50  & 19.27 $\pm$ 0.45  & 19.17 $\pm$ 0.42  \\ \hline      griewank & 10  & 15.18 $\pm$ 12.21  & \bf{12.82 $\pm$ 9.30}  & 15.07 $\pm$ 9.69  & 20.56 $\pm$ 12.21  \\ \hline      griewank & 20  & 27.34 $\pm$ 12.03  & \bf{18.84 $\pm$ 5.62}  & 20.12 $\pm$ 7.59  & 37.70 $\pm$ 19.87  \\ \hline      griewank & 30  & 39.46 $\pm$ 13.21  & \bf{31.33 $\pm$ 12.98}  & 35.29 $\pm$ 9.84  & 50.00 $\pm$ 15.22  \\ \hline      rastrigin & 10  & \bf{73.63 $\pm$ 16.69}  & 79.21 $\pm$ 18.00  & 84.20 $\pm$ 17.47  & 79.99 $\pm$ 18.15  \\ \hline      rastrigin & 20  & \bf{117.38 $\pm$ 30.23}  & 152.84 $\pm$ 34.60  & 138.08 $\pm$ 23.11  & 155.38 $\pm$ 31.97  \\ \hline      rastrigin & 30  & \bf{143.47 $\pm$ 29.35}  & 176.62 $\pm$ 26.82  & 175.89 $\pm$ 26.73  & 191.12 $\pm$ 40.51  \\ \hline     \multicolumn{2}{|c|}{Average-Ranking}  & 2.00  & 1.93  & 2.60  & 3.47  \\ \hline    \multicolumn{2}{|c|}{Wilcoxon-Test} & Control method  & 4-9-2  & 5-8-2  & 9-6-0  \\ \hline \multicolumn{2}{|c|}{Wilcoxon-Test}  & 2-9-4  & Control method & 0-15-0  & 7-8-0  \\ \hline \end{tabular}
\end{table*}

To further analyse the efficacy of two proposed portfolio frameworks, we compare their performance with other two frameworks: random selection (RS) policy and epsilon-greedy policy with $\epsilon=0.5$ (EG50). RS policy randomly selects an algorithm at each generation while EG50 policy selects the ``optimal'' algorithm with probability 0.5 according to the cumulative reward and selects a random algorithm with probability 0.5 \cite{peter2014coco}. The comparative results on test problems are presented in Table \ref{result2}.

As shown in the result table, either UCB-IBSAEA or Par-IBSAEA obtain the best result among four frameworks except in 10-dimension ackley problem that RS gets a relatively better solution. Furtherly, the Par-IBSAEA performs better in sphere and griewank problem which are regarded as uni-modal problems. And UCB-IBSAEA is always the best framework among four frameworks in rosenbrock and rastrigin problems which are complex multi-modal problems. And the four frameworks also have the similar performance in Ackley function.

Compared with UCB-IBSAEA, RS, EG50, Par-IBSAEA is the only parallel framework that running algorithm candidate in parallel. It is better than other three sequential frameworks in sphere and griewank problems. As discussed in the above subsection, algorithm portfolio increases the diversity during the optimization process by combining different kinds of algorithms. Par-IBSAEA uses all algorithm simultaneously at each generation, using the same database that algorithm candidates search the next re-evaluated solution from various aspects. This idea is similar to the negative correlation search (NCS) \cite{ke2016negative}, which uses different agents to cover different regions of search space. However, if the number of algorithm candidates increases, the performance of Par-IBSAEA might deteriorate because the limited computational cost cannot afford the huge cost for parallelly running many algorithms simultaneously at each iteration.

\begin{figure}
  \centering
  \includegraphics[scale=0.45]{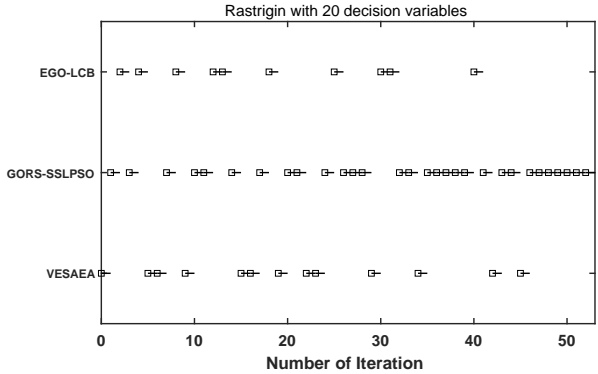}
  \caption{An illustration of selection behaviour for UCB-IBSAEA in 20-dimension rastrigin problem.} 
  \label{choice}
\end{figure}

Although Par-IBSAEA is outstanding in uni-modal problems, UCB-IBSAEA is superior to other frameworks in multi-modal problems as shown in Table \ref{result2}. It selects the appropriate algorithm at each generation according to the UCB policy in Eq. \eqref{ucb-t}. The GORS-SSLPSO is the best algorithm among three algorithm candidates and UCB-IBSAEA is likely to detect the fact and allocate more computational budget to this algorithm. Take the 20-dimension rastrigin problem as an example, the behaviour of selection in each iteration of UCB-IBSAEA is plotted in Figure \ref{choice}. We could find UCB-IBSAEA explores three algorithms at the early stage and then detect that GORS-SSLPSO is better than another two algorithms so that it mostly selects GORS-SSLPSO at the later stage.

On the other hand, we find that there is no big difference between random selection policy and two proposed frameworks in Table \ref{result2}. This is mainly because there are only three algorithms in our experiments that the best algorithm will be selected at a high probability. Furtherly, the randomness could provide diversity for portfolio framework and avoid framework being trapped in one action so that RS policy might obtain the best result than other frameworks, like 10-dimension ackley problem. Meanwhile, the difference among algorithm candidates is not very large so that random selection will not perform very badly.

\subsection{Risk analysis}
The risk of proposed framework for individual-based SAEAs will be analyzed in this subsection. The metric to measure algorithms' risk is employed from the work in \cite{peng2010population}, in which the comparative risk is estimated by comparing the quality of solutions obtained by two algorithms in a set of test problems with several runs. Considering benchmark problems $\mathcal{F} = \{f_k|k=1, 2, ..., n\}$ and algorithm constituents $\mathcal{A} = \{A_j|j=1,2, ..., m\}$, the probability of $A_i$ outperforming $A_j$ as $A_i \succ A_j$ can be calculated by the following equation: 
\begin{equation}
  P(A_i \succ A_j) = \frac{1}{n}\sum_{k=1}^{n}P(q_{i, k} < q_{j, k} | f_k)
\end{equation}
where the $q_{i, k}$ denotes the quality of solution obtained by $A_i$ in $f_k$. And $P(q_{i, k} < q_{j, k} | f_k)$ can be estimated by the following equation:
\begin{equation}
  P(q_{i, k} < q_{j, k} | f_k) = \frac{\sum_{s=1}^{s_i} \sum_{t=1}^{s_j}  \mathcal{I} (y_{i, k, s} < y_{j, k, t}) }{s_i \times s_j} 
\end{equation}
where $y_{i, k, s}$ represents the fitness of solution obtained by $A_i$ in $s_{th}$ trial for $f_k$ problem, $s_i$, $s_j$ represent the number of trails of each algorithm for one problem and $\mathcal{I}(\cdot)$ denotes the indicator function. 

\begin{table}[htpb!]
  \caption{The comparison risk of algorithm portfolio and single algorithm. The two figures in each cell stand for the probabilities that the portfolio and the individual-based SAEAs  outperformed each other.}\label{risk}
  \begin{tabular}{|c|c|c|c|}
    \hline
               & GORS-SSLPSO   & VESAEA        & EGO-LCB     \\ \hline
    UCB-IBSAEA & 0.60 - 0.40   & 0.75 - 0.25   & 0.94 - 0.06 \\ \hline
    Par-IBSAEA & 0.55 - 0.45   & 0.72 - 0.28   & 0.94 - 0.06 \\ \hline
    RS         & 0.56 - 0.44   & 0.71 - 0.29   & 0.94 - 0.06 \\ \hline
    EG50       & 0.55 - 0.45   & 0.68 - 0.32   & 0.94 - 0.06 \\ \hline
    \end{tabular}
  \end{table}

The comparative risk between portfolio frameworks and three SAEAs are shown in the Table \ref{risk}. The four portfolio frameworks are all better than single SAEAs, in which the UCB-IBSAEA is slightly better than other three frameworks. Firstly, we can conclude that the portfolio framework could obviously reduce the risk of failing in optimizing problems. The performance of SAEAs could be promoted by combining various efficient algorithms so that the CEPs could be solved much better within limited computational resources. Secondly, we can find the proposed UCB-IBSAEA is more effective than other three frameworks in terms of optimisation risk so that the UCB-IBSAEA is more appropriate to be used for an unknown CEP.  

\section{Conclusion} \label{conclusion}

\doublecheck{In this paper, the algorithm portfolio for individual-based SAEAs is the first time to be considered in solving computationally very expensive black-box problems. We creatively proposed two effective portfolio frameworks as the Par-IBSAEA and UCB-IBSAEA framework. The Par-IBSAEA runs all algorithms in parallel and all algorithm candidates share the same database at each iteration. On the other hand, the UCB-IBSAEA considers the portfolio as a multi-armed bandit problem and employs the UCB-Tuned algorithm to formulate the selection policy. Meanwhile, we design a new efficient reward definition for optimization in the UCB-IBSAEA framework, which normalizes the quality of solution in the scope of a sliding window. Additionally, the numerical experiments are performed to analyze the performance of the proposed frameworks in five commonly used test problems with dimension $d=10, 20, 30$. The results shows that the Par-IBSAEA can increase the diversity of optimization process and gets a much better performance than all the algorithm constituents in uni-modal problems as well as serveral multi-model problems even though a significant terrible algorithm is included in the framework, and the UCB-IBSAEA can effectively balance the exploration and exploitation in algorithm selection, which obtains an excellent performance than other frameworks in multi-modal problems. }

This work mainly focuses on individual-based SAEAs only for solving CEPs. In the future, the portfolio framework will be improved to suit for more general optimisation algorithms. On the other hand, it is much important to select appropriate algorithm candidates for algorithm portfolio which directly influence the framework's performance, so it is valuable to do more research on how to choose constituent algorithms for algorithm portfolio in SAEAs.

\section*{Acknowledgments}
This work was supported by the National Key R$\&$D Program of China (Grant No. 2017YFC0804003), the Program for Guangdong Introducing Innovative and Enterpreneurial Teams (Grant No. 2017ZT07X386), Shenzhen Peacock Plan (Grant No. KQTD2016112514355531), the Science and Technology Innovation Committee Foundation of Shenzhen (Grant No. ZDSYS201703031748284, Grant No. JCYJ20180504165652917 and Grant No. JCYJ20170817112421757) and the Program for University Key Laboratory of Guangdong Province (Grant No. 2017KSYS008).

\bibliographystyle{ACM-Reference-Format}
\bibliography{reference} 


\begin{thebibliography}{28}


\ifx \showCODEN    \undefined \def \showCODEN     #1{\unskip}     \fi
\ifx \showDOI      \undefined \def \showDOI       #1{#1}\fi
\ifx \showISBNx    \undefined \def \showISBNx     #1{\unskip}     \fi
\ifx \showISBNxiii \undefined \def \showISBNxiii  #1{\unskip}     \fi
\ifx \showISSN     \undefined \def \showISSN      #1{\unskip}     \fi
\ifx \showLCCN     \undefined \def \showLCCN      #1{\unskip}     \fi
\ifx \shownote     \undefined \def \shownote      #1{#1}          \fi
\ifx \showarticletitle \undefined \def \showarticletitle #1{#1}   \fi
\ifx \showURL      \undefined \def \showURL       {\relax}        \fi
\providecommand\bibfield[2]{#2}
\providecommand\bibinfo[2]{#2}
\providecommand\natexlab[1]{#1}
\providecommand\showeprint[2][]{arXiv:#2}

\bibitem[\protect\citeauthoryear{Auer, Cesa{-}Bianchi, and Fischer}{Auer
  et~al\mbox{.}}{2002}]%
        {peter2002finite}
\bibfield{author}{\bibinfo{person}{Peter Auer}, \bibinfo{person}{Nicol{\`{o}}
  Cesa{-}Bianchi}, {and} \bibinfo{person}{Paul Fischer}.}
  \bibinfo{year}{2002}\natexlab{}.
\newblock \showarticletitle{Finite-time Analysis of the Multiarmed Bandit
  Problem}.
\newblock \bibinfo{journal}{{\em Machine Learning\/}} \bibinfo{volume}{47},
  \bibinfo{number}{2-3} (\bibinfo{year}{2002}), \bibinfo{pages}{235--256}.
\newblock


\bibitem[\protect\citeauthoryear{Baudis}{Baudis}{2014}]%
        {peter2014coco}
\bibfield{author}{\bibinfo{person}{Petr Baudis}.}
  \bibinfo{year}{2014}\natexlab{}.
\newblock \showarticletitle{COCOpf: An Algorithm Portfolio Framework}.
\newblock \bibinfo{journal}{{\em CoRR\/}}  \bibinfo{volume}{abs/1405.3487}
  (\bibinfo{year}{2014}).
\newblock


\bibitem[\protect\citeauthoryear{Baudi{\v{s}} and
  Po{\v{s}}{\'\i}k}{Baudi{\v{s}} and Po{\v{s}}{\'\i}k}{2014}]%
        {baudivs2014online}
\bibfield{author}{\bibinfo{person}{Petr Baudi{\v{s}}} {and}
  \bibinfo{person}{Petr Po{\v{s}}{\'\i}k}.} \bibinfo{year}{2014}\natexlab{}.
\newblock \showarticletitle{Online black-box algorithm portfolios for
  continuous optimization}. In \bibinfo{booktitle}{{\em International
  Conference on Parallel Problem Solving from Nature}}. Springer,
  \bibinfo{pages}{40--49}.
\newblock


\bibitem[\protect\citeauthoryear{Cauwet, Liu, Rozi{\`e}re, and Teytaud}{Cauwet
  et~al\mbox{.}}{2016}]%
        {cauwet2016algorithm}
\bibfield{author}{\bibinfo{person}{Marie-Liesse Cauwet},
  \bibinfo{person}{Jialin Liu}, \bibinfo{person}{Baptiste Rozi{\`e}re}, {and}
  \bibinfo{person}{Olivier Teytaud}.} \bibinfo{year}{2016}\natexlab{}.
\newblock \showarticletitle{Algorithm portfolios for noisy optimization}.
\newblock \bibinfo{journal}{{\em Annals of Mathematics and Artificial
  Intelligence\/}} \bibinfo{volume}{76}, \bibinfo{number}{1-2}
  (\bibinfo{year}{2016}), \bibinfo{pages}{143--172}.
\newblock


\bibitem[\protect\citeauthoryear{Cauwet, Liu, and Teytaud}{Cauwet
  et~al\mbox{.}}{2014}]%
        {cauwet2014algorithm}
\bibfield{author}{\bibinfo{person}{Marie-Liesse Cauwet},
  \bibinfo{person}{Jialin Liu}, {and} \bibinfo{person}{Olivier Teytaud}.}
  \bibinfo{year}{2014}\natexlab{}.
\newblock \showarticletitle{Algorithm portfolios for noisy optimization:
  Compare solvers early}. In \bibinfo{booktitle}{{\em International Conference
  on Learning and Intelligent Optimization}}. Springer, \bibinfo{pages}{1--15}.
\newblock


\bibitem[\protect\citeauthoryear{Elsayed, Sarker, Essam, and Hamza}{Elsayed
  et~al\mbox{.}}{2014}]%
        {saber2014testing}
\bibfield{author}{\bibinfo{person}{Saber~M. Elsayed}, \bibinfo{person}{Ruhul~A.
  Sarker}, \bibinfo{person}{Daryl~Leslie Essam}, {and} \bibinfo{person}{Noha~M.
  Hamza}.} \bibinfo{year}{2014}\natexlab{}.
\newblock \showarticletitle{Testing united multi-operator evolutionary
  algorithms on the {CEC2014} real-parameter numerical optimization}. In
  \bibinfo{booktitle}{{\em Proceedings of the {IEEE} Congress on Evolutionary
  Computation, {CEC} 2014, Beijing, China, July 6-11, 2014}}. IEEE,
  \bibinfo{pages}{1650--1657}.
\newblock


\bibitem[\protect\citeauthoryear{Fialho, Schoenauer, and Sebag}{Fialho
  et~al\mbox{.}}{2010}]%
        {fialho2010toward}
\bibfield{author}{\bibinfo{person}{{\'{A}}lvaro Fialho}, \bibinfo{person}{Marc
  Schoenauer}, {and} \bibinfo{person}{Mich{\`{e}}le Sebag}.}
  \bibinfo{year}{2010}\natexlab{}.
\newblock \showarticletitle{Toward comparison-based adaptive operator
  selection}. In \bibinfo{booktitle}{{\em Genetic and Evolutionary Computation
  Conference, {GECCO} 2010, Proceedings, Portland, Oregon, USA, July 7-11,
  2010}}. \bibinfo{pages}{767--774}.
\newblock


\bibitem[\protect\citeauthoryear{Hoffman, Brochu, and de~Freitas}{Hoffman
  et~al\mbox{.}}{2011}]%
        {hoffman2011portfolio}
\bibfield{author}{\bibinfo{person}{Matthew~D Hoffman}, \bibinfo{person}{Eric
  Brochu}, {and} \bibinfo{person}{Nando de Freitas}.}
  \bibinfo{year}{2011}\natexlab{}.
\newblock \showarticletitle{Portfolio Allocation for Bayesian Optimization.}.
  In \bibinfo{booktitle}{{\em UAI}}. Citeseer, \bibinfo{pages}{327--336}.
\newblock


\bibitem[\protect\citeauthoryear{Huberman, Lukose, and Hogg}{Huberman
  et~al\mbox{.}}{1997}]%
        {huberman1997economics}
\bibfield{author}{\bibinfo{person}{Bernardo~A Huberman},
  \bibinfo{person}{Rajan~M Lukose}, {and} \bibinfo{person}{Tad Hogg}.}
  \bibinfo{year}{1997}\natexlab{}.
\newblock \showarticletitle{An economics approach to hard computational
  problems}.
\newblock \bibinfo{journal}{{\em Science\/}} \bibinfo{volume}{275},
  \bibinfo{number}{5296} (\bibinfo{year}{1997}), \bibinfo{pages}{51--54}.
\newblock


\bibitem[\protect\citeauthoryear{Jin}{Jin}{2005}]%
        {jin2005comprehensive}
\bibfield{author}{\bibinfo{person}{Yaochu Jin}.}
  \bibinfo{year}{2005}\natexlab{}.
\newblock \showarticletitle{A comprehensive survey of fitness approximation in
  evolutionary computation}.
\newblock \bibinfo{journal}{{\em Soft Computing\/}} \bibinfo{volume}{9},
  \bibinfo{number}{1} (\bibinfo{year}{2005}), \bibinfo{pages}{3--12}.
\newblock


\bibitem[\protect\citeauthoryear{Jin}{Jin}{2011}]%
        {jin2011surrogate}
\bibfield{author}{\bibinfo{person}{Yaochu Jin}.}
  \bibinfo{year}{2011}\natexlab{}.
\newblock \showarticletitle{Surrogate-assisted evolutionary computation: Recent
  advances and future challenges}.
\newblock \bibinfo{journal}{{\em Swarm and Evolutionary Computation\/}}
  \bibinfo{volume}{1}, \bibinfo{number}{2} (\bibinfo{year}{2011}),
  \bibinfo{pages}{61--70}.
\newblock


\bibitem[\protect\citeauthoryear{Jin and Sendhoff}{Jin and Sendhoff}{2009}]%
        {jin2009systems}
\bibfield{author}{\bibinfo{person}{Yaochu Jin} {and} \bibinfo{person}{Bernhard
  Sendhoff}.} \bibinfo{year}{2009}\natexlab{}.
\newblock \showarticletitle{A systems approach to evolutionary multiobjective
  structural optimization and beyond}.
\newblock \bibinfo{journal}{{\em IEEE Computational Intelligence Magazine\/}}
  \bibinfo{volume}{4}, \bibinfo{number}{3} (\bibinfo{year}{2009}).
\newblock


\bibitem[\protect\citeauthoryear{Jin, Wang, Tinkle, Dan, and Kiettinen}{Jin
  et~al\mbox{.}}{2018}]%
        {jin2018data}
\bibfield{author}{\bibinfo{person}{Yaochu Jin}, \bibinfo{person}{Handing Wang},
  \bibinfo{person}{Chugh Tinkle}, \bibinfo{person}{Guo Dan}, {and}
  \bibinfo{person}{Miettinen Kiettinen}.} \bibinfo{year}{2018}\natexlab{}.
\newblock \showarticletitle{Data-Driven Evolutionary Optimization: An Overview
  and Case Studies}.
\newblock \bibinfo{journal}{{\em IEEE Transactions on Evolutionary
  Computation\/}} (\bibinfo{year}{2018}), \bibinfo{pages}{1--1}.
\newblock
\showISSN{1089-778X}
\showDOI{%
\url{https://doi.org/10.1109/TEVC.2018.2869001}}


\bibitem[\protect\citeauthoryear{Jones, Schonlau, and Welch}{Jones
  et~al\mbox{.}}{1998}]%
        {jones1998efficient}
\bibfield{author}{\bibinfo{person}{Donald~R Jones}, \bibinfo{person}{Matthias
  Schonlau}, {and} \bibinfo{person}{William~J Welch}.}
  \bibinfo{year}{1998}\natexlab{}.
\newblock \showarticletitle{Efficient global optimization of expensive
  black-box functions}.
\newblock \bibinfo{journal}{{\em Journal of Global optimization\/}}
  \bibinfo{volume}{13}, \bibinfo{number}{4} (\bibinfo{year}{1998}),
  \bibinfo{pages}{455--492}.
\newblock


\bibitem[\protect\citeauthoryear{Katehakis and Jr.}{Katehakis and Jr.}{1987}]%
        {michael1987bandit}
\bibfield{author}{\bibinfo{person}{Michael~N. Katehakis} {and}
  \bibinfo{person}{Arthur F.~Veinott Jr.}} \bibinfo{year}{1987}\natexlab{}.
\newblock \showarticletitle{The Multi-Armed Bandit Problem: Decomposition and
  Computation}.
\newblock \bibinfo{journal}{{\em Math. Oper. Res.\/}} \bibinfo{volume}{12},
  \bibinfo{number}{2} (\bibinfo{year}{1987}), \bibinfo{pages}{262--268}.
\newblock


\bibitem[\protect\citeauthoryear{Peng, Tang, Chen, and Yao}{Peng
  et~al\mbox{.}}{2010}]%
        {peng2010population}
\bibfield{author}{\bibinfo{person}{Fei Peng}, \bibinfo{person}{Ke Tang},
  \bibinfo{person}{Guoliang Chen}, {and} \bibinfo{person}{Xin Yao}.}
  \bibinfo{year}{2010}\natexlab{}.
\newblock \showarticletitle{Population-based algorithm portfolios for numerical
  optimization}.
\newblock \bibinfo{journal}{{\em IEEE Transactions on evolutionary
  computation\/}} \bibinfo{volume}{14}, \bibinfo{number}{5}
  (\bibinfo{year}{2010}), \bibinfo{pages}{782--800}.
\newblock


\bibitem[\protect\citeauthoryear{Shahriari, Swersky, Wang, Adams, and
  De~Freitas}{Shahriari et~al\mbox{.}}{2016}]%
        {shahriari2016taking}
\bibfield{author}{\bibinfo{person}{Bobak Shahriari}, \bibinfo{person}{Kevin
  Swersky}, \bibinfo{person}{Ziyu Wang}, \bibinfo{person}{Ryan~P Adams}, {and}
  \bibinfo{person}{Nando De~Freitas}.} \bibinfo{year}{2016}\natexlab{}.
\newblock \showarticletitle{Taking the human out of the loop: A review of
  bayesian optimization}.
\newblock \bibinfo{journal}{{\it Proc. IEEE}} \bibinfo{volume}{104},
  \bibinfo{number}{1} (\bibinfo{year}{2016}), \bibinfo{pages}{148--175}.
\newblock


\bibitem[\protect\citeauthoryear{St{-}Pierre and Liu}{St{-}Pierre and
  Liu}{2014}]%
        {david2014differential}
\bibfield{author}{\bibinfo{person}{David~Lupien St{-}Pierre} {and}
  \bibinfo{person}{Jialin Liu}.} \bibinfo{year}{2014}\natexlab{}.
\newblock \showarticletitle{Differential Evolution algorithm applied to
  non-stationary bandit problem}. In \bibinfo{booktitle}{{\em Proceedings of
  the {IEEE} Congress on Evolutionary Computation, {CEC} 2014, Beijing, China,
  July 6-11, 2014}}. IEEE, \bibinfo{pages}{2397--2403}.
\newblock


\bibitem[\protect\citeauthoryear{Tang, Yang, and Yao}{Tang
  et~al\mbox{.}}{2016}]%
        {ke2016negative}
\bibfield{author}{\bibinfo{person}{Ke Tang}, \bibinfo{person}{Peng Yang}, {and}
  \bibinfo{person}{Xin Yao}.} \bibinfo{year}{2016}\natexlab{}.
\newblock \showarticletitle{Negatively Correlated Search}.
\newblock \bibinfo{journal}{{\em {IEEE} Journal on Selected Areas in
  Communications\/}} \bibinfo{volume}{34}, \bibinfo{number}{3}
  (\bibinfo{year}{2016}), \bibinfo{pages}{542--550}.
\newblock


\bibitem[\protect\citeauthoryear{Tong, Huang, Liu, and Yao}{Tong
  et~al\mbox{.}}{2019}]%
        {hao2018voronoi}
\bibfield{author}{\bibinfo{person}{Hao Tong}, \bibinfo{person}{Changwu Huang},
  \bibinfo{person}{Jialin Liu}, {and} \bibinfo{person}{Xin Yao}.}
  \bibinfo{year}{2019}\natexlab{}.
\newblock \bibinfo{title}{Voronoi-based Efficient Surrogate-assisted
  Evolutionary Algorithm for Very Expensive Problems}.
\newblock   (\bibinfo{year}{2019}).
\newblock
\showeprint{arXiv:1901.05755}


\bibitem[\protect\citeauthoryear{Vrugt, Robinson, and Hyman}{Vrugt
  et~al\mbox{.}}{2009}]%
        {jasper2009self}
\bibfield{author}{\bibinfo{person}{Jasper~A. Vrugt}, \bibinfo{person}{Bruce~A.
  Robinson}, {and} \bibinfo{person}{James~M. Hyman}.}
  \bibinfo{year}{2009}\natexlab{}.
\newblock \showarticletitle{Self-Adaptive Multimethod Search for Global
  Optimization in Real-Parameter Spaces}.
\newblock \bibinfo{journal}{{\em IEEE Transactions on Evolutionary
  Computation\/}} \bibinfo{volume}{13}, \bibinfo{number}{2}
  (\bibinfo{year}{2009}), \bibinfo{pages}{243--259}.
\newblock


\bibitem[\protect\citeauthoryear{Wang, Jin, and Doherty}{Wang
  et~al\mbox{.}}{2017}]%
        {wang2017committee}
\bibfield{author}{\bibinfo{person}{Handing Wang}, \bibinfo{person}{Yaochu Jin},
  {and} \bibinfo{person}{John Doherty}.} \bibinfo{year}{2017}\natexlab{}.
\newblock \showarticletitle{Committee-based active learning for
  surrogate-assisted particle swarm optimization of expensive problems}.
\newblock \bibinfo{journal}{{\em IEEE Transactions on Cybernetics\/}}
  \bibinfo{volume}{47}, \bibinfo{number}{9} (\bibinfo{year}{2017}),
  \bibinfo{pages}{2664--2677}.
\newblock


\bibitem[\protect\citeauthoryear{Wang, Jin, and Jansen}{Wang
  et~al\mbox{.}}{2016}]%
        {wang2016data}
\bibfield{author}{\bibinfo{person}{Handing Wang}, \bibinfo{person}{Yaochu Jin},
  {and} \bibinfo{person}{Jan~O Jansen}.} \bibinfo{year}{2016}\natexlab{}.
\newblock \showarticletitle{Data-driven surrogate-assisted multiobjective
  evolutionary optimization of a trauma system}.
\newblock \bibinfo{journal}{{\em IEEE Transactions on Evolutionary
  Computation\/}} \bibinfo{volume}{20}, \bibinfo{number}{6}
  (\bibinfo{year}{2016}), \bibinfo{pages}{939--952}.
\newblock


\bibitem[\protect\citeauthoryear{Wolpert and Macready}{Wolpert and
  Macready}{1997}]%
        {wolpert1997no}
\bibfield{author}{\bibinfo{person}{David~H Wolpert} {and}
  \bibinfo{person}{William~G Macready}.} \bibinfo{year}{1997}\natexlab{}.
\newblock \showarticletitle{No free lunch theorems for optimization}.
\newblock \bibinfo{journal}{{\em IEEE transactions on evolutionary
  computation\/}} \bibinfo{volume}{1}, \bibinfo{number}{1}
  (\bibinfo{year}{1997}), \bibinfo{pages}{67--82}.
\newblock


\bibitem[\protect\citeauthoryear{Xu, Hutter, Hoos, and Leyton-Brown}{Xu
  et~al\mbox{.}}{2008}]%
        {xu2008satzilla}
\bibfield{author}{\bibinfo{person}{Lin Xu}, \bibinfo{person}{Frank Hutter},
  \bibinfo{person}{Holger~H Hoos}, {and} \bibinfo{person}{Kevin Leyton-Brown}.}
  \bibinfo{year}{2008}\natexlab{}.
\newblock \showarticletitle{SATzilla: portfolio-based algorithm selection for
  SAT}.
\newblock \bibinfo{journal}{{\em Journal of artificial intelligence
  research\/}}  \bibinfo{volume}{32} (\bibinfo{year}{2008}),
  \bibinfo{pages}{565--606}.
\newblock


\bibitem[\protect\citeauthoryear{Yu, Tan, Sun, and Zeng}{Yu
  et~al\mbox{.}}{2019}]%
        {yu2019generation}
\bibfield{author}{\bibinfo{person}{Haibo Yu}, \bibinfo{person}{Ying Tan},
  \bibinfo{person}{Chaoli Sun}, {and} \bibinfo{person}{Jianchao Zeng}.}
  \bibinfo{year}{2019}\natexlab{}.
\newblock \showarticletitle{A generation-based optimal restart strategy for
  surrogate-assisted social learning particle swarm optimization}.
\newblock \bibinfo{journal}{{\em Knowledge-Based Systems\/}}
  \bibinfo{volume}{163} (\bibinfo{year}{2019}), \bibinfo{pages}{14--25}.
\newblock


\bibitem[\protect\citeauthoryear{Yuen, Chow, Zhang, and Lou}{Yuen
  et~al\mbox{.}}{2016}]%
        {yuen2016algorithm}
\bibfield{author}{\bibinfo{person}{Shiu~Yin Yuen}, \bibinfo{person}{Chi~Kin
  Chow}, \bibinfo{person}{Xin Zhang}, {and} \bibinfo{person}{Yang Lou}.}
  \bibinfo{year}{2016}\natexlab{}.
\newblock \showarticletitle{Which algorithm should I choose: an evolutionary
  algorithm portfolio approach}.
\newblock \bibinfo{journal}{{\em Applied Soft Computing\/}}
  \bibinfo{volume}{40} (\bibinfo{year}{2016}), \bibinfo{pages}{654--673}.
\newblock


\bibitem[\protect\citeauthoryear{Zhang, Georgiopoulos, and
  Anagnostopoulos}{Zhang et~al\mbox{.}}{2014}]%
        {tian2014online}
\bibfield{author}{\bibinfo{person}{Tiantian Zhang}, \bibinfo{person}{Michael
  Georgiopoulos}, {and} \bibinfo{person}{Georgios~C. Anagnostopoulos}.}
  \bibinfo{year}{2014}\natexlab{}.
\newblock \showarticletitle{Online model racing based on extreme performance}.
  In \bibinfo{booktitle}{{\em Genetic and Evolutionary Computation Conference,
  {GECCO} '14, Vancouver, BC, Canada, July 12-16, 2014}}. ACM,
  \bibinfo{pages}{1351--1358}.
\newblock


\end{thebibliography}

\end{document}